\documentclass[submission,copyright,creativecommons]{eptcs}
\providecommand{\event}{UITP 2012} 
\usepackage{tikz}
\usetikzlibrary{decorations.markings}
\usetikzlibrary{arrows}
\usepackage{amssymb}
\usepackage{amsfonts,amsmath,latexsym,stmaryrd}
\usepackage{listings}

\newcommand{\C}[1]{\mbox{\lstinline`#1`}}

\definecolor{dkblue}{rgb}{0,0.1,0.5} 
\definecolor{lightblue}{rgb}{0,0.5,0.5} 
\definecolor{dkgreen}{rgb}{0,0.4,0} 
\definecolor{dk2green}{rgb}{0.4,0,0} 
\definecolor{dkviolet}{rgb}{0.6,0,0.8}
\definecolor{shadethmcolor}{rgb}{0.9, 0.9,1}

\usepackage{theorem}
\usepackage{rotating}
\usepackage{pdflscape}
\usepackage{cite}
\usepackage{etex, xy}
\xyoption{all}

\theorembodyfont{\rmfamily}

\makeatletter

\let\c@table\c@figure
\makeatother

\newtheorem{Exmp}[figure]{Example}

\title{Machine Learning in Proof General: Interfacing Interfaces}
\author{Ekaterina Komendantskaya\footnote{The work was supported by EPSRC grant EP/J014222/1.}
\institute{School of Computing\\
University of Dundee, UK}
\email{katya@computing.dundee.ac.uk}
\and
J\'onathan Heras$^\ast$
\institute{School of Computing\\
University of Dundee, UK}
\email{jonathanheras@computing.dundee.ac.uk}
\and
Gudmund Grov\footnote{The work was supported by EPSRC grants EP/H023852/1 and EP/H024204/1.}
\institute{School of Mathematical \& Computer Sciences\\ 
Heriot-Watt University, UK}
\email{G.Grov@hw.ac.uk}
}
\def\titlerunning{Machine Learning in Proof General: Interfacing Interfaces}
\def\authorrunning{E. Komendantskaya, J. Heras \& G. Grov}

\begin{document}
\maketitle

\begin{abstract}
\noindent\textbf{Abstract:} We present ML4PG --- a machine learning extension for Proof General.
It allows users to gather proof statistics related to shapes of goals, sequences of applied tactics, and proof tree structures from the libraries of interactive higher-order proofs 
written in Coq and SSReflect.
The gathered data is clustered using the state-of-the-art machine learning algorithms available in MATLAB and Weka.
ML4PG provides automated interfacing between Proof General and MATLAB/Weka.
The results of clustering are used by ML4PG to provide proof hints in the process of interactive proof development.  

\end{abstract}

\section{Introduction}\label{sec:introduction}

Over the last few decades, theorem proving has seen major developments.
\emph{Automated (first-order) theorem provers (ATPs)} (e.g. E~\cite{E}, Vampire~\cite{vampire} and SPASS~\cite{spass}) and SAT/SMT solvers
(e.g. CVC3~\cite{cvc3}, Yices~\cite{yices} and Z3~\cite{z3})
are becoming increasingly fast and efficient~\cite{Kuehlwein2012}. \emph{Interactive (higher-order) theorem provers (ITPs)} (e.g. Coq~\cite{Coq},
Isabelle/HOL~\cite{NPW02}, Agda~\cite{Agda}, Matita~\cite{Matita} and Mizar~\cite{mizar}) have been enriched with dependent types, 
(co)inductive types, type classes and now provide rich programming environments~\cite{FCT, FOOT, flyspeck, KMM00}.

The main conceptual difference between  ATPs and ITPs lies in the styles of proof development: for ATPs, the proof process is primarily an automatically performed \emph{proof search},
for ITPs it is mainly user-driven \emph{proof development}.
Nevertheless, ITPs have seen major advances in proof 
automation~\cite{SSReflect,HT98,sledgehammer}. One particular trend is to re-enforce proof automation in ITPs by employing state-of-the-art 
tools from ATPs~\cite{Laurent1,sledgehammer}, SAT/SMT solvers~\cite{SledgehammerSMT,Laurent1,SATACL2} or Computer Algebra systems~\cite{HT98,Analytica,GAPCOQ}.
One major success of this approach is Sledgehammer~\cite{Paulson_threeyears}: it offers Isabelle/HOL users an option to call for an ATP/SMT-generated solution~\cite{SledgehammerSMT}.

Integrating ITPs with ATPs requires a lot of research into \emph{methods of interfacing}. 
Namely, the major challenge is a \emph{sound} and \emph{reliable} translation between inherently different languages~\cite{MP09,HT98,GAPCOQ,Laurent1,TACAS13}. 
This especially concerns interpreting outputs from ATPs back into the higher-order environment~\cite{Laurent1,MP09,TACAS13}, 
which we will also call here \emph{backward interfacing}. For example, Sledgehammer uses the results provided by external tools to guide the higher-order proofs, but leaves it to the 
Isabelle/HOL kernel to check that the suggested tactic combination is valid. 

In parallel to the work mentioned above, another trend of research has been developed. 
It approaches the issue of improving proof automation from the perspective of statistical
and machine learning methods.
Several  aspects of automated and interactive theorem proving can be data-mined:

\begin{itemize}
  \item proof heuristics can be data-mined to improve proof search in ATPs~\cite{DFGS99,DenzingerS00,APACL2,KL12,KuhlweinLTUH12,TsivtsivadzeUGH11,UrbanSPV08,Urban06}; 
  \item history of successful and unsuccessful proof attempts can be used to inform interactive proof development in ITPs~\cite{Duncan02,KL12}.
\end{itemize}
 
The former trend has been more successful, mainly directed to improve premise selection in ATPs. 
In the case of
higher-order interactive proofs, there are four main issues that make statistical data-mining challenging:
\begin{itemize}
\item[\textbf{C.1.}] The richer tactic language reduces the chance of finding regularities and proof patterns. 
ITP-based proofs involve
an unlimited variety of structures and proof patterns, in comparison to ATPs, where resolution or rewriting may be the two possible rules to apply.
Hence, finding statistically significant proof features becomes challenging.
    
\item[\textbf{C.2.}]  The notions of a  \emph{proof} may be regarded from different perspectives in ITPs: it may be seen as a transition between
the subgoals \cite{DenzingerS00,UrbanSPV08,Urban06},
a combination of applied tactics \cite{Duncan02}, or --- more traditionally --- a proof tree showing the overall proof strategy  \cite{KL12}. 
Depending on the nature of the proof and application areas of the machine learning tools, each of the three aspects can be important for statistical proof pattern recognition.

\item[\textbf{C.3.}]  Backward interfacing --- interpreting results provided by the statistical machine learning tool back into the higher-order interactive prover --- can be a challenge.

\item[\textbf{C.4.}]  In interactive proofs, the most time-consuming and challenging part is no longer the time the prover takes to find the proof.
It is the time the proof developer takes to understand and guide the proof. Therefore, when data-mining interactive proofs,
we are interested not only in the final result --- the successful proof, but also in the \emph{proof process}, including failed and discarded derivation steps.
We want machine learning to guide the process, not to diagnose or speed up already found proofs.
For this, machine learning tools for ITPs need to be interactive.
\end{itemize} 

Up to now, experiments on data-mining interactive proofs were always constrained by
the lack of the interactive interfacing between machine learning algorithms and the user-driven proof development. 
For example, in~\cite{APACL2}, there was a tool that gathered statistics, but no automated data-mining tools were used;
in~\cite{KL12}, there was a feature extraction method to data-mine proofs  but it was not connected to efficient statistics gathering;
in~\cite{Duncan02}, these two were semi-automated. 

Because of  the inherently  interactive nature of proofs in ITPs, user interfaces for ITPs play an important role in proof development.
For our experiments, we chose \emph{Proof General}~\cite{ProofGeneral} --- a general-purpose, emacs-based interface for a range of higher-order
theorem provers, e.g. Isabelle, Coq or Lego. Among them, we have chosen Coq~\cite{Coq} and 
its SSReflect library~\cite{SSReflect} for our experiments. Although both built upon the same language --- \emph{Calculus of Inductive Constructions} \cite{CoquandH88}, they have distinct proof styles,
analysis of which plays a special role in this paper, see Section \ref{sec:alternative}. 

This idea of maintaining a strong, convenient interface for a range of proof systems
is mirrored by a similar trend in the machine learning community. As statistical methods require 
users to constantly interpret and monitor results 
computed by the statistical tools, the community has developed  uniform
interfaces --- convenient environments in which the user can choose which machine learning algorithm to use for processing the data and for interpreting results.
One such famous interface we take for our experiments is MATLAB~\cite{Matlab} which has its own underlying
programming language, and comprises several machine learning toolboxes, from general-purpose \emph{Statistical Toolbox} to the specialised \emph{Neural-network Toolbox}.
The second major machine learning interface we explore is Weka~\cite{Weka} 
-- an open source, general purpose interface to run a wide variety of machine learning algorithms.

We have already referred to the two different meanings of the
term ``interfaces''.
On the one hand, interfacing may mean translation mechanisms connecting ITPs with other proof automation tools \cite{MP09,HT98,GAPCOQ,Laurent1}; and on the other hand, it 
is used as a synonym for user-friendly environment. In this paper, the two views on the notion of interfaces meet.
Our primary goal is to integrate the state-of-the-art machine learning technology into ITPs, in order to 
improve user experience and productivity. However, since machine learning algorithms will need to gather statistics from the user's
behaviour, and feed the results back to the user \emph{during} the proof development process,
this primary task will never be accomplished without machine learning becoming an integral part of the user interface. 
 
\begin{figure}
\begin{center}
 \begin{tikzpicture}
\draw[fill=gray,draw=gray] (-1.95,.45) rectangle (2.05,-.55);  
\draw[fill=white] (-2,.5) rectangle (2,-.5); 
\node (0,0) {Proof General interface};

\draw[fill=gray,draw=gray] (8.05,.45) rectangle (12.05,-.55);  
\draw[fill=white] (8,.5) rectangle (12,-.5); 
\draw (10,0) node{MATLAB, Weka};

\draw (5,0) node{\textbf{ML4PG}};

\draw[latex-,shorten <=2pt,shorten >=2pt,dashed] (8,.4) .. controls (6,1) and (4,1) .. (2,.4); 
\draw (5,1.1) node[anchor=north,fill=white]{\emph{\small{feature extraction}}};
 
\draw[latex-,shorten <=2pt,shorten >=2pt,dashed] (2,-.4) .. controls (4,-1) and (6,-1) .. (8,-.4); 
\draw (5,-1.1) node[anchor=south,fill=white]{\emph{\small{proof families}}};

\draw[fill=white,rounded corners] (-2,-2) rectangle (2,-3); 
\draw (0,-2.3) node{Interactive Prover:} ;
\draw (0,-2.7) node{Coq, SSReflect} ;

\draw[dotted,shorten <=2pt,shorten >=2pt,-latex] (0,-.5) -- (0,-2);
\draw[dotted,shorten <=2pt,shorten >=2pt,-latex] (10,-.5) -- (10,-2); 
 
\draw[fill=white,rounded corners] (8,-2) rectangle (12,-3); 
\draw (10,-2.3) node{Clustering algorithms:} ;
\draw (10,-2.7) node{k-means, Gaussian, \ldots} ;
\end{tikzpicture}
\end{center}
\caption{\emph{\footnotesize{\textbf{``Interfacing Interfaces'' with ML4PG.}}}}\label{fig:II}

\end{figure}
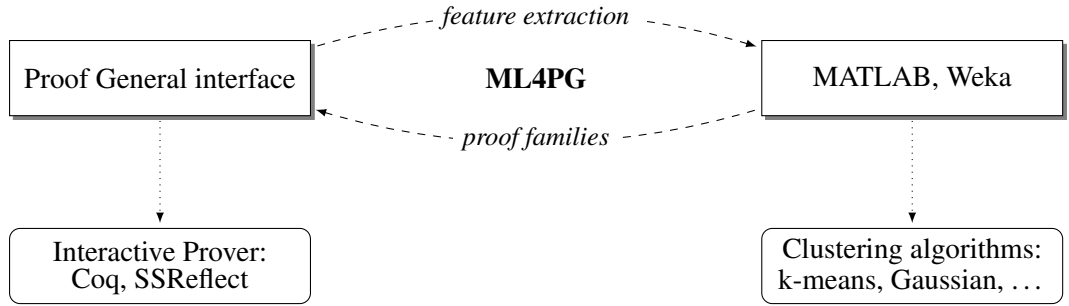

In this paper, we show the results of our work on \emph{interfacing interfaces} --- building a user-friendly environment that
integrates a range of machine learning tools provided by MATLAB and Weka into Proof General. In particular, we pay attention to addressing the challenges \textbf{C.1-C.4}. 
We implement the following vision of interfacing between ITP and machine learning, and call the result 
ML4PG (\emph{machine learning for Proof General}), see Figure \ref{fig:II}.\footnote{It is available at \cite{HK12}, where the reader can download ML4PG, user manual 
and examples (see also~\cite{CICM13}).}

\begin{enumerate}
\item  ML4PG must be able to gather statistics from interactive proofs (challenge \textbf{C.3}), and relate this statistics accurately to
the three aspects of ITP-based proof development: goal-level, tactic-level, and proof tree level (challenge \textbf{C.2}). 
We focus on this issue in Section~\ref{sec:levels}.

\item ML4PG must automatically extract the relevant features associated with these three aspects in a form 
suitable for machine learning tools --- that is, numerical vectors of fixed length, also known as \emph{feature vectors} (challenges \textbf{C.1--C.2}).
We present a new method of \emph{proof-trace} feature extraction in Section~\ref{sec:alternative}.

\item ML4PG must enable the user to choose from a range of machine learning interfaces and algorithms suitable for proof data-mining (challenge \textbf{C.4}).
As we do not assume the Proof General user to have machine learning expertise, we want to delegate a substantial amount of pre- and post-processing of statistical
results to ML4PG. Section~\ref{sec:clustering} deals with these questions.

\item ML4PG must automatically connect to the chosen machine learning interface, and it should collect,  appropriately analyse and interpret the output of these algorithms, at any stage of the interactive proof (challenge \textbf{C.3}). 
Note that in our work ``backward interfacing'' from the machine learning tools to Proof General 
is less demanding compared to~\cite{MP09,HT98,GAPCOQ,Laurent1}. We do not seek a translation of statistical results into the Coq \emph{language}; instead, we use the
statistical results to inform the user of arising proof patterns during the proof development. 
As Sections~\ref{sec:clustering} and~\ref{sec:session} show, this kind of light backward interfacing can be efficiently implemented.

\item Finally, ML4PG must interact with  the user by providing relevant information about the user's current proof goal in relation 
to statistically similar proof patterns detected in different libraries or even across different users (challenges \textbf{C.1--C.4}).
We discuss this in Sections~\ref{sec:clustering} and~\ref{sec:session}.  
\end{enumerate}

There are two aspects to this work: development of methods of interactive interfacing between the ITP and machine learning interfaces; and a more general aspect of studying the potential of machine-learning 
methods in proof-pattern recognition. 
This paper mainly focuses on the the first aspect. 
References \cite{CICM13,HK12,KuhlweinLTUH12} are specifically devoted to the benchmarks, evaluation and discussion of statistical proof-pattern
recognition methods in theorem proving.  
As far as ML4PG interface engineering goes, it was important for us to make accessible a number of simple but useful options that the user with no experience in machine learning could use.
Section~\ref{sec:rw} surveys related work on integration of machine learning with theorem proving.
Finally, in Section \ref{sec:conclusions}, we conclude and discuss future extensions.

\section{The Three Levels of an Interactive Proof}\label{sec:levels}

In this section, we consider a variety of possible approaches to proof pattern recognition in ITPs; namely, we consider automated proofs from the levels
of goal transitions, tactic sequences, and proof trees. 

We start with several running examples to illustrate the kind of statistical help we expect from ML4PG.
We consider the library containing various lemmas about natural numbers and lists.

\begin{Exmp}\label{ex:example0}
Suppose the user starts with the following two lemmas about multiplication by $0$: \texttt{Lemma mult\_n\_0:}$\forall n:nat, 0 = n * 0$ and 
\texttt{Lemma mult\_0\_n}: $\forall n:nat, 0 = 0 * n$, see left side of Tables \ref{tab:E12} and \ref{tab:E5}
for their proofs.
They state two very similar properties, however, the proofs for them are different; notably, one proof involved induction, while another involved only simplification. 

\begin{table*}
	\centering
	\footnotesize{
		\begin{tabular}{|l|l|}
		\hline
		Goals & Tactics \\
		\hline
		\hline
		$\forall n:nat, 0 = n * 0$ & \\
			& \lstinline?induction n.? \\
		1. $ 0 = 0 * 0$; 2. $0 = S n * 0$ & \\
& \lstinline?simpl; trivial.?   \\
    	 $0 = S n * 0$ & \\
& \lstinline?simpl; trivial.?   \\
				$\Box$ & \\
		& \lstinline?Qed.?\\
		\hline
		\end{tabular}
				\begin{tabular}{|l|l|}
		\hline
		Goals & Tactics \\
		\hline
		\hline
	$\forall l: list~A, l ++ [] = l$& \\
			& \lstinline?induction l.? \\
			1.  $[] ++ [] = []$.  2. $(a :: l) ++ [] = a :: l$ & \\
		& \lstinline?simpl;trivial.? \\		
		$(a :: l) ++ [] = a :: l$ & \\
		& \lstinline?simpl.? \\
		$a :: l ++ [] = a :: l$ & \\
		& \lstinline?rewrite IHl.? \\
		$a :: l = a :: l$ & \\
		& \lstinline?trivial.? \\
		$\Box$ & \\
		& \lstinline?Qed.?\\
		\hline
		\end{tabular}
				
		}
	\caption{\footnotesize{\emph{\textbf{Proof steps for Lemmas  $\texttt{mult\_n\_0}: \forall n:nat, 0 = n * 0$ and \texttt{app\_l\_nil}: $\forall l: list~A, l ++ [] = l$.}}}}
	\label{tab:E12}
\end{table*}

Next, suppose the user switches to the library containing lists; and needs some guidance to proceed with the proofs for
\texttt{Lemma app\_l\_nil}: $\forall l: list~A, l ++ [] = l$,
and
  \texttt{Lemma app\_nil\_l}: $\forall l: list~A, [] ++ l = l$.
	The user asks ML4PG  to ``statistically match'' these problems to previously seen proofs in the same or in  a different library. 
	We then want ML4PG to tell the user that there are two similar lemmas in the \texttt{Nat} library --- namely \texttt{Lemma mult\_n\_0:}$\forall n:nat, 0 = n * 0$ and 
\texttt{Lemma mult\_0\_n}: $\forall n:nat, 0 = 0 * n$. Then the user will adapt these old proofs to complete new proofs as given on the right side of Tables~\ref{tab:E12} and~\ref{tab:E5}.
Note that this guidance will go further than just identifying proofs over the same data type, identifying same tactic combinations, same functions/operations or similar lemma shapes.
Such guidance would be based on statistical correlation of several proof features.

\begin{table*}
	\centering
	\footnotesize{
	\begin{tabular}{|l|l|}
		\hline
		Goals & Tactics \\
		\hline
		\hline
		$\forall n:nat, 0 = 0 * n$ & \\
			& \lstinline?intro.? \\
		 $0 =  0 * n$ & \\
& \lstinline?simpl; trivial.?   \\
				$\Box$ & \\
		& \lstinline?Qed.?\\
		\hline
		\end{tabular}
		\begin{tabular}{|l|l|}
		\hline
		Goals & Tactics \\
		\hline
		\hline
		$\forall l: list~A, [] ++ l = l$ & \\
		& \lstinline?intro l? \\
		$[] ++ l = l$ & \\
		& \lstinline?simpl;trivial.?\\
		$\Box$ & \\
		& \lstinline?Qed.? \\
		\hline
		\end{tabular}}
	\caption{\footnotesize{\emph{\textbf{Proof steps for Lemmas  $\texttt{mult\_0\_n}: \forall n:nat, 0 = 0 * n$ and \texttt{app\_nil\_l}: $\forall l: list~A, [] ++ l = l$.}}}}
	\label{tab:E5}
\end{table*}

As can be seen from these examples, the user may be interested in 
data-mining the proofs based on either 

\begin{itemize}
\item[\textbf{D.1.}] transitions between subgoal-shapes (in which case Lemmas \texttt{app\_l\_nil} and \texttt{mult\_n\_0} of Table \ref{tab:E12} have common patterns), or 
\item[\textbf{D.2.}] statistics of tactic combination (in which case Lemmas \texttt{app\_nil\_l} and \texttt{mult\_0\_n} of Table \ref{tab:E5} should be identified), or 
\item[\textbf{D.3.}] a more general understanding of lemma content (in which case all four are similar).
\end{itemize}

\end{Exmp}

Therefore, we distinguish three levels at which pattern-recognition in ITP proofs can be approached, see also ~\cite{GKA12}:

\begin{enumerate}
\item \emph{Goal-pattern recognition}. 
Sequences of subgoals may show an apparent pattern in the structure of the
formulas. This type of feature
abstraction has been used for learning the inputs for automatic
provers~\cite{UrbanSPV08,DenzingerS00} --- which has later been
extended to interactive proofs \cite{TsivtsivadzeUGH11}. 

\begin{Exmp}
The left-most columns of Tables~\ref{tab:E12} and~\ref{tab:E5} should be used to gather such information about goal-patterns.
\end{Exmp}

\item \emph{Tactic-pattern recognition}. Sequences of tactics applied at every level of the proof
bear some apparent patterns, as well. There is always a finite number
of tactics for any given proof, and therefore, they can serve as
features for statistical learning.   Previous work on
learning proof strategies \cite{Jam02,Duncan02,JDB11} has taken this
approach.  It is important to note that there
may be proofs in which the goal structures do not bear any evident
pattern; however, the sequence of applied tactics does. Also, as an
additional complication, there is a variety of tactic
combinations that may lead to a successful proof for one goal;
and conversely, different goals may yield same sequences of
tactics in successful proofs. Moreover, tactics often have complex
configurations, which can be hidden or given as arguments (e.g.
rules to apply or instantiations of variables). 

\begin{Exmp}
The right-most columns of Tables~\ref{tab:E12} and~\ref{tab:E5} provide information about such tactic-patterns.
\end{Exmp}

The disadvantage of tactic-pattern recognition is
that any knowledge of when and why a tactic is applied, as well as
its result is lost (except with respect to other tactic
applications).

\item \emph{Proof tree pattern recognition}. Finally, there is the level of a proof tree that shows relations
between different proof branches and subgoals and gives a better view
of the overall proof flow; this approach was tested in \cite{KL12}
using multi-layer neural networks and kernels. 

\begin{figure}
 
\begin{center}
\footnotesize{
\begin{tikzpicture}
\coordinate (0) at (0,0);
\coordinate (1) at (0,-1);
\coordinate (2_1) at (-2,-2);
\coordinate (3_1) at (-2,-3);
\coordinate (2_2) at (5,-2);
\coordinate (3_2) at (5,-3);
\coordinate (4_2) at (5,-4);
\coordinate (5_2) at (5,-5);

\begin{scope}[decoration={
    markings,
    mark=at position 0.6 with {\arrow{*}}}
    ] 

\draw[-*] (1) -- (0,-1.5) node[anchor=south west]{{\tiny induction l}};
\draw (0,-1.4) -- (2_1);
\draw (0,-1.4) -- (2_2);
\draw[postaction={decorate}] (2_1) -- node[anchor=west]{{\tiny simpl; trivial}} (3_1);
\draw[postaction={decorate}] (2_2) -- node[anchor=west]{{\tiny simpl}} (3_2);
\draw[postaction={decorate}] (3_2) -- node[anchor=west]{{\tiny rewrite IHl}}(4_2);
\draw[postaction={decorate}] (4_2) -- node[anchor=west]{{\tiny trivial}} (5_2);
\draw (1) node[fill=white] {\verb"forall l: list A, l ++ [] = l"}; 
\draw (2_1) node[fill=white]  {\verb" [] ++ [] = []"}; 
\draw (2_2) node[fill=white]  {\verb" (a: A) (l: list A) (IHl: l ++ [] = l) |- (a::l) ++ [] = a::l"}; 
\draw (3_1) node[fill=white]  {$\square$};
\draw (3_2) node[fill=white]  {\verb"(a: A) (l: list A) (IHl: l ++ [] = l) |- a::l ++ [] = a::l"}; 
\draw (4_2) node[fill=white]  {\verb"(a: A) (l: list A) (IHl: l ++ [] = l) |- a::l = a::l"}; 
\draw (5_2) node[fill=white]  {$\square$};
\end{scope}

\end{tikzpicture}}

\end{center}
\caption{\footnotesize{\emph{\textbf{Proof tree for \texttt{app\_l\_nil}.}}}}\label{fig:E1tree}

\end{figure}

\begin{Exmp}
Figure \ref{fig:E1tree} shows the proof tree for \texttt{app\_l\_nil}.
An advantage of the proof tree as opposed to goal or tactic sequence, is that it distinguishes between different proof branches.

\end{Exmp}

 \end{enumerate}
 
Our second running example is based on  
the \lstinline?bigop? library of SSReflect. This library is devoted to generic indexed big operations, like $\sum\limits_{i=0}^n f(i)$ or $\bigcup\limits_{i\in I} f(i)$.

\begin{Exmp}
We take three lemmas about number series: 
$$\forall n, 2(\sum\limits^n_{i=0} i) = n(n+1); \ \ \ \forall n, \sum\limits^{2n}_{i=0|odd~i} i = n^2; \ \ \  \forall n, \prod\limits^n_1 i = n!$$ 
The proofs of these three lemmas, both at the level of goals and tactics, are given in Table~\ref{tab:sumfirstn}.
Intuitively, they show certain similarities and dissimilarities, both at the level of goals and tactics.
In the next sections, we will test how ML4PG analyses such cases. 
\end{Exmp}

 \begin{table*}
 	\centering
 	\footnotesize{
 		\begin{tabular}{|l|l|}
 		\hline
 		Goal & Tactic \\
 		\hline
 		\hline
 	$2(\sum\limits^n_{i=0} i) = n(n+1)$ & \\
 			& {\scriptsize \lstinline?elim : n.?} \\
        $2(\sum\limits^0_{i=0} i) = 0 \times 1$ & \\
        & {\scriptsize\lstinline?by rewrite mul0n big_nat1 muln0.?}\\
        $\forall n, 2(\sum\limits^n_{i=0} i) = n(n+1) \implies 2(\sum\limits^{n+1}_{i=0} i) = (n+1)(n+2)$ &\\
        & {\scriptsize\lstinline?move => n IH.?} \\
        $2(\sum\limits^{n+1}_{i=0} i) = (n+1)(n+2)$ &\\
        &{\scriptsize\lstinline?by rewrite big_nat_recr mulnDr IH -mulnDl addn2 ?}\\
        &~~~~~~{\scriptsize\lstinline?mulnC.?}\\
 		$\Box$ & \\
 		& {\scriptsize\lstinline?Qed.?}\\
 		\hline
 		\end{tabular}
 		
 		\begin{tabular}{|l|l|}
 		\hline
 		Goal & Tactic \\
 		\hline
 		\hline
 	$\sum\limits^{2n}_{i=0|odd~i} i = n^2$ & \\
 			& {\scriptsize\lstinline?elim : n.?} \\
        $\sum\limits^{2\times 0}_{i=0|odd~i} i = 0^2$ & \\
        & {\scriptsize\lstinline?rewrite exp0n // /index_iota subn0 big1_seq //.?}\\
        $\forall i \in \mathbb{N}, odd~i~\&\&~(i \in iota~0~(2 \times 0)) \implies i = 0$ & \\
        &{\scriptsize \lstinline?by move => i; move/andP => [_ H2]; move : H2;?}\\
        & ~~~~~~{\scriptsize\lstinline?rewrite muln0 in_nil.?}\\
        
        $\forall n, \sum\limits^{2n}_{i=0|odd~i} i = n^2 \implies \sum\limits^{2(n+1)}_{i=0|odd~i} i = (n+1)^2$ &\\
        & \lstinline?move => n IH.? \\
        $\sum\limits^{2(n+1)}_{i=0|odd~i} i = (n+1)^2$ &\\
        &{\scriptsize\lstinline?by rewrite big_mkcond -[n.+1]addn1 mulnDr muln1 ?}\\
        &~~~~~~{\scriptsize\lstinline?addn2 !big_nat_recr IH odd2n odd2n1 //= addn0 ?}\\
        &~~~~~~{\scriptsize\lstinline?n1square n2square.?}\\
 		$\Box$ & \\
 		& \lstinline?Qed.?\\
 		\hline
 		\end{tabular}
 		
 		\begin{tabular}{|l|l|}
 		\hline
 		Goal & Tactic \\
 		\hline
 		\hline
 	$\prod\limits^n_1 i = n!$ & \\
 			& {\scriptsize\lstinline?elim : n.?} \\
        $\prod_1^1 i = 0!$ & \\
        & {\scriptsize\lstinline?by rewrite big_nil.?}\\
        $\forall n, \prod\limits^n_1 i = n! \implies \prod\limits^{n+1}_1 i = (n+1)!$ &\\
        & {\scriptsize\lstinline?move => n IH.?} \\
        $\prod\limits^{n+1}_1 i = (n+1)!$ &\\
        &{\scriptsize\lstinline?by rewrite factS big_add1 -IH big_add1 big_nat_recr mulnC.?}\\
 		$\Box$ & \\
 		& \lstinline?Qed.?\\
 		\hline
 		\end{tabular}
 		 		}
 	\caption{\footnotesize{\emph{\textbf{Interactive Proofs for \texttt{Lemma sum\_first\_n}: $2(\sum\limits^n_{i=0} i) = n(n+1)$; \texttt{Lemma sum\_first\_n\_odd}:
 	$\sum\limits^{2n}_{i=0|odd~i} i = n^2$; and \texttt{Lemma fact\_prod}: $\prod\limits^n_1 i = n!$.}}}}
 	\label{tab:sumfirstn}
 \end{table*}

Next, we study how these general considerations about the levels of proof patterns are used in ML4PG to extract 
features used in statistical data-mining.

\section{Feature Extraction in ML4PG: the Proof Trace Method}\label{sec:alternative}

In this section, we explain algorithms used by ML4PG to gather proof statistics
at the levels of goals, tactics, and proof trees. 

The discovery of statistically significant features in data is a research area of its own in machine learning, known as 
\emph{feature extraction}, see~\cite{Bishop}.
Irrespective of the particular feature-extraction algorithm used, most pattern-recognition tools will require that the number of selected features is limited and fixed.\footnote{\emph{``Sparse methods''} of machine-learning is an exception to this rule, see \cite{KuhlweinLTUH12,Mash}. We discuss the issue in Section \ref{sec:rw}.} 
We design our own method of proof feature extraction. 
The major challenge is to respect the above restriction while allowing to data-mine potentially unlimited variety of different higher-order formulas and proofs. 

We first focus on the level of goals. 
 ML4PG must choose the relevant features for statistical analysis. 
At this level, we could consider general goal  properties such as ``goal shape'' (e.g. ``associative-shape'' or ``commutative-shape''),  or properties like ``the goal
embeds a hypothesis'', \lq\lq{}the goal is embedded into a hypothesis\rq\rq{}. 
However, gathering such features uniformly across any set of higher-order proofs  would be hard, 
especially when working with richer theories and dependent types.

\begin{Exmp}
Consider the proof for \texttt{app\_l\_nil} given in Table~\ref{tab:E5}. One could say that the valuable information about the shape of the (sub)goal 4 is that it embeds 
the inductive hypothesis, 
as this fact is later used in the proof.
However, for more complex examples, deciding such embeddings unambiguously during 
feature-extraction may be difficult, see \cite{BB05}.
Finally, detecting a fixed number of properties like e.g. ``commutativity'' may apply to one type of proof
libraries, e.g. natural numbers, but not to others, e.g. lists, in which case uniform comparison of proof patterns across libraries becomes hard. 
\end{Exmp}


\begin{table}
 \centering
\footnotesize{
 \begin{tabular}{|l||l|l|l|l|l|l|}
 \hline
  & \emph{tactics} & \emph{N tactics} & \emph{arg type} & \emph{tactic arg is hypothesis?} & \emph{top symbol} & \emph{n subgoals} \\
 \hline
 \hline
 \emph{g1} & \textbf{induction} & $\mathbf{1}$ & $\mathbf{nat}$  & \textbf{no} & \textbf{forall} & \textbf{2} \\
 \hline
   \emph{g2}& \textbf{simpl;trivial} & $\mathbf{2}$ & \textbf{none} & \textbf{no} & \textbf{equal} & \textbf{0} \\
  \hline
   \emph{g3} & \verb"simpl;trivial" & $2$ & \textbf{none}  & \textbf{no} & \textbf{equal} & 0 \\
  \hline
 \emph{g4} & - & - & - & - & - & - \\
  \hline
 \emph{g5} & - &  - & - & - & - & - \\
  \hline 
   \end{tabular}
 	
 \begin{tabular}{|l||l|l|l|l|l|l|}
 \hline
  & \emph{tactics} & \emph{N tactics} & \emph{arg type} & \emph{tactic arg is hypothesis?} & \emph{top symbol} & \emph{n subgoals} \\
 \hline
 \hline
 \emph{g1} & \textbf{induction} & $\mathbf{1}$ & $\mathbf{list}$  & \textbf{no} & \textbf{forall} & \textbf{2} \\
 \hline
   \emph{g2}& \textbf{simpl;trivial} & $\mathbf{2}$ & \textbf{none} & \textbf{no} & \textbf{equal} & \textbf{0} \\
  \hline
  \emph{g3} & \verb"simpl" & $1$ & \textbf{none}  & \textbf{no} & \textbf{equal} & 1 \\
  \hline
   \emph{g4} & \verb"rewrite" & $1$ & Prop & IH & \lstinline?equal? & 1 \\
  \hline
  \emph{g5} & \verb"trivial" &  $1$ & \textbf{none} & \textbf{no} & \lstinline?equal? & \textbf{0} \\
  \hline 
   \end{tabular}}
  \caption{\footnotesize{\emph{\textbf{Goal-level feature extraction tables.} Parameters inside the double lines are the extracted features. Notation \emph{g1--g5} is used to denote 
  five consecutive
subgoals in the derivation. Columns are the properties of subgoals the feature extraction method of ML4PG will track: names of applied tactics, their number, arguments, link of the proof
step to a hypothesis (Hyp), inductive hypothesis (IH) or a library lemma; top symbol of the current goal, and the number of the generated subgoals.}
\emph{The two tables show a comparison of the goal-level feature tables for  \lstinline?mult_n_0? and \lstinline?app_l_nil?. Intuitively, 18 out of 30 features in these tables 
show correlation; we highlight them in bold.}}}\label{tab:listnil}
  \end{table}

This is why we developed a method of implicit tracking of proof properties, called the \emph{proof trace method}; its early variant was
used in \cite{KL12}. The idea is as follows. When direct feature-extraction of the goal shapes is infeasible, 
we still can infer some properties of the goals when gathering statistics of how the user treats the goals.
In other words, we let the term-structure show itself
through the proof steps it induces. 
We deliberately do not pre-define the types of \emph{proof patterns} that ML4PG must recognise, or define what a \emph{correlation of proof features}
is. Instead, we want statistical machine learning tools to suggest the user what these might be.   
An advantage of such proof feature extraction method is that it applies uniformly to any Coq library, and does not require any adaptation when ML4PG changes the libraries.

Another important feature ML4PG must be sensitive to, is the long-lasting effect of one proof step 
on several consecutive proof steps.
The dependency between subgoals very often extends much farther than from one proof step to its immediate successor. 
Thus, we want to capture two dimensions of goal transformation in a proof:

\begin{enumerate}
	\item various traceable properties of a single (sub)-goal;
	\item transformations of each such property throughout several proof steps.
\end{enumerate}

This is why we 
design two dimensional arrays as shown in Table~\ref{tab:listnil}
to allow for statistical data-mining of the two dimensions in their relation.

\begin{Exmp}
Consider Table \ref{tab:listnil} where the correlation between \lstinline?mult_n_0? and \lstinline?app_l_nil? at the goal level is shown. If we consider the table associated with
\lstinline?app_l_nil?, 
the fact of using the tactics \verb"induction" and \verb"(simpl;trivial)" may not be significant, as this combination can be applied to a variety of goals. 
It may be insignificant, on its own, that the top symbol of the goal was the quantifier $\forall$. 
However, the table related to this lemma allows us to characterise the goal $\forall \texttt{ l : list A, []++l=l}$ by the 30 features (entries) of the table.
Correlations of values of these features will be more likely to show significant proof patterns, if such exist.

\end{Exmp}

\begin{Exmp}
As can be seen in Table~\ref{tab:series_goal}, there is a strong correlation between the features associated with 
the first step in the proofs of \lstinline?sum_first_n?, \lstinline?sum_first_n_odd? and \lstinline?fact_prod?.
However, this strong correlation only remains between \lstinline?sum_first_n? and \lstinline?fact_prod?
when successive proof steps are considered. This illustrates the fact that we cannot focus just on the first goal of a proof, but we
have to study its proof trace to obtain relevant patterns.

\begin{table}
\centering
\footnotesize{
\begin{tabular}{|l||l|l|l|l|l|l|}
\hline
 & \emph{tactics} & \emph{N tactics} & \emph{arg type} & \emph{arg is hyp?} & \emph{top symbol} & \emph{n subgoals} \\
\hline
\hline
\emph{g1}& elim & $1$  & nat  & Hyp & equal & $2$ \\
 \hline
  \emph{g2} & rewrite & $1$  & $3\times$Prop  & EL1 & equal & $0$ \\
 \hline
  \emph{g3} & move =$>$ & $1$  & nat,Prop& no & forall & $1$ \\
 \hline
  \emph{g4} & rewrite & $1$  & $6\times$Prop & EL2  & equal & $0$ \\
 \hline 
 \emph{g5} & - & - & - & - & - & - \\
 \hline
  \end{tabular}

\begin{tabular}{|l||l|l|l|l|l|l|}
\hline
 & \emph{tactics} & \emph{N tactics} & \emph{arg type} & \emph{arg is hyp?} & \emph{top symbol} & \emph{n subgoals} \\
\hline
\hline
\emph{g1}& \textbf{elim} & $\mathbf{1}$  & \textbf{nat}  & \textbf{Hyp} & \textbf{equal} & $\mathbf{2}$ \\
\hline
\emph{g2}& \textbf{rewrite} & $\mathbf{1}$  & $4\times$Prop  & EL3 & \textbf{equal} & $1$ \\
 \hline
  \emph{g3} & move=$>$,move/ & $4$ & nat, $5\times$Prop  & EL4 & forall & $0$ \\
    &  move:, rewrite &  & & & &  \\
 \hline
  \emph{g4} & move =$>$ & $\mathbf{1}$  & nat,Prop & no & forall & $1$ \\
 \hline
  \emph{g5} & rewrite & $1$  & $11\times$Prop & EL5 & equal & $\mathbf{0}$ \\
 \hline 
  \end{tabular}
  
\begin{tabular}{|l||l|l|l|l|l|l|}
\hline
 & \emph{tactics} & \emph{N tactics} & \emph{arg type} & \emph{arg is hyp?} & \emph{top symbol} & \emph{n subgoals} \\
\hline
\hline
\emph{g1}& \textbf{elim} & $\mathbf{1}$  & \textbf{nat}  & \textbf{Hyp} & \textbf{equal} & $\mathbf{2}$ \\
 \hline
  \emph{g2} & \textbf{rewrite} & $\mathbf{1}$  & Prop  & big\_nil & \textbf{equal} & $\mathbf{0}$ \\
 \hline
  \emph{g3} & \textbf{move =$>$} & $\mathbf{1}$  & \textbf{nat,Prop} & \textbf{no} & \textbf{forall} & $\mathbf{1}$ \\
 \hline
  \emph{g4} & \textbf{rewrite} & $\mathbf{1}$  & $\mathbf{6\times}$\textbf{Prop} & EL6 & \textbf{equal} & $\mathbf{0}$\\
 \hline 
 \emph{g5} & \textbf{-} &  \textbf{-} & \textbf{-} & \textbf{-} & \textbf{-} & \textbf{-} \\
 \hline 
  \end{tabular}}
 \caption{\footnotesize{\emph{\textbf{Goal-level feature extraction tables for lemmas \lstinline?sum_first_n?, \lstinline?sum_first_n_odd? and \lstinline?fact_prod?. }
  In the tables of \lstinline?sum_first_n_odd? and \lstinline?fact_prod?.
  We use the following notation to note when ML4PG gathers the lemma names:  $EL1$ stands for (\lstinline?mul0n?, \lstinline?big_nat1? and \lstinline?muln0?),
  $EL2$ for (\lstinline?big_nat_recr?, \lstinline?mulnDr?, \lstinline?IH?, \lstinline?mulnD1?, \lstinline?addn2? and \lstinline?mulnC?),
  $EL3$  for  (\lstinline?exp0n?, \lstinline?index_iota? \lstinline?subn0? and \lstinline?big1_seq?),
   $EL4$ for  (\lstinline?/andP?, \lstinline?muln0? and \lstinline?in_nil?),  
   $EL5$ for  (\lstinline?big_mkcond?, \lstinline?addn1?, \lstinline?mulnDr?, \lstinline?muln1?, \lstinline?addn2?, \lstinline?big_nat_recr?, \lstinline?IH?,
   \lstinline?odd2n?, \lstinline?odd2n1?, \lstinline?addn0?, \lstinline?n1square? and \lstinline?n2square?) 
   and  $EL6$ for  (\lstinline?factS?, \lstinline?big_add1?, \lstinline?IH?, \lstinline?big_add1?, \lstinline?big_nat_recr? and \lstinline?mulnC?);
   the lemmas and libraries can be found in \cite{HK12}.  We highlight in bold the correlation with the features of \lstinline?sum_first_n?.
  There is a strong correlation between \lstinline?sum_first_n? and \lstinline?fact_prod? (27 out of 30 features); on the contrary, there is a weak
  correlation between \lstinline?sum_first_n? and \lstinline?sum_first_n_odd? (11 out of 30 features). }}}\label{tab:series_goal}
 \end{table}
\end{Exmp}

Another advantage of the method is its focus on user interaction:
 ML4PG learns proof patterns specific to the user's proof style as given in the chosen library of proofs. 




On the tactic level, ML4PG focuses on features associated with each tactic applied in a proof script. 
The action of the tactic-level feature extraction algorithm is shown in  Table~\ref{tab:E1T}.
It is worth noting that the structure of the goal-level Table~\ref{tab:listnil} can be reused in all the systems based on the application of
a sequence of tactics (e.g. Coq, Isabelle/HOL, Matita, etc.); the only difference would be the values which 
populate the table. On the other hand, the structure of the tactic-level table depends on the concrete system, since each ITP
has its concrete set of tactics.


The case of Coq is special, since we can find two proof styles: plain Coq and SSReflect. Although SSReflect is an extension of Coq, 
this package implements a set of 
proof tactics designed to support the extensive use of small-scale reflection in formal proofs~\cite{SSReflect}. 
In addition, the behaviour of some Coq tactics has been modified (for instance, the \lstinline?rewrite? tactic); 
so, the SSReflect imposes a distinct proof style. ML4PG works with both styles of Coq
proofs. In the case of plain Coq, the rows of the tactic table represent the main Coq tactics (from almost $100$ Coq tactics
ML4PG currently distinguishes $10$ most popular). The set of SSReflect tactics consists of less than $10$ tactics, so we have included all of them. 

\begin{Exmp}

Consider the fragments of tactic-level tables associated with the Lemma \texttt{app\_nil\_l} and \texttt{mult\_O\_n}, in Table~\ref{tab:E1T}.
The extracted features show close correlation, as expected.

\begin{table}
\centering
\footnotesize{
\begin{tabular}{|l||c|c|c|c|c|}
\hline
 & \emph{arg-1 type} & \emph{rest arg types} & \emph{arg is hyp? }& \emph{top symbol} & \emph{n times used} \\
\hline
\hline
 \lstinline?intro? & list & \textbf{none} & \textbf{no}  & \textbf{forall} & \textbf{1 }\\
\hline
 \lstinline?case? & \textbf{-}&\textbf{-}&\textbf{-}&\textbf{-}&\textbf{-}\\
\hline
 \lstinline?simpl? & \textbf{none}&\textbf{none}&\textbf{no} &\textbf{equal}&\textbf{1}\\
\hline
 \lstinline?trivial? & \textbf{none}&\textbf{none}&\textbf{no} &\textbf{equal}&\textbf{1}\\
\hline

 \end{tabular}
\begin{tabular}{|l||c|c|c|c|c|}
\hline
 & \emph{arg-1 type} & \emph{rest arg types} & \emph{arg is hyp? }& \emph{top symbol} & \emph{n times used} \\
\hline
\hline
 \lstinline?intro? & nat & \textbf{none} & \textbf{no}  & \textbf{forall} & \textbf{1} \\
\hline
 \lstinline?case? & \textbf{-}&\textbf{-}&\textbf{-}&\textbf{-}&\textbf{-}\\
\hline
 \lstinline?simpl? & \textbf{none}&\textbf{none}&\textbf{no} &\textbf{equal}&\textbf{1}\\
\hline
 \lstinline?trivial? & \textbf{none}&\textbf{none}&\textbf{no} &\textbf{equal}&\textbf{1}\\
\hline
 \end{tabular}}
\caption{\footnotesize{\emph{
\textbf{Tactic-level feature extraction table for Coq.} The rows are the tactics implemented in the given ITP.
The columns of this table encode the same properties for both 
plain Coq proofs and proofs in SSReflect: the type of the first argument, a number that denotes the types of the remaining tactic arguments; 
identification whether the arguments of the tactic are hypotheses (Hyp), inductive hypotheses (IH), external lemmas or none of them.
Finally, ML4PG populates the last two columns with the list of top symbols of the goals where the tactic has been applied; and the number of times the tactic was applied.
Tactic-level feature extraction tables for \texttt{mult\_O\_n} and \texttt{app\_nil\_l}  show close correlation (in bold).}}}
\label{tab:E1T}
\end{table}
 
\end{Exmp}

\begin{Exmp}

 Fragments of tactic tables for Lemmas \lstinline?sum_first_n?, \lstinline?sum_first_n_odd? and \lstinline?fact_prod?
 are given in Table~\ref{tab:mult_tactic}. There is a strong correlation between \lstinline?sum_first_n? and \lstinline?fact_prod?
 at this level.

\begin{table}
\centering
\footnotesize{
\begin{tabular}{|l||l|l|l|l|l|}
\hline
 & \emph{arg-1 type} & \emph{rest arg types} & \emph{arg is hyp? }& \emph{top symbol} & \emph{n times used} \\
\hline
\hline
 \lstinline?move =>? & nat & Prop& None & forall& 1 \\
\hline
 \lstinline?move : ? & -&- &- &- &- \\
\hline
 \lstinline?move/ ? & -&- &- &- &- \\
\hline
 \lstinline?rewrite? & Prop & $8\times$Prop& EL'  &$2\times$ equal&2\\
\hline 
\lstinline?case? & - & -&- &- &-  \\
  \hline
 \lstinline?elim? &nat &- & Hyp & equal  & 1\\
 \hline

  \end{tabular}
  
  \begin{tabular}{|l||l|l|l|l|l|}
\hline
 & \emph{arg-1 type} & \emph{rest arg types} & \emph{arg is hyp? }& \emph{top symbol} & \emph{n times used} \\
\hline
\hline
 \lstinline?move =>? &\textbf{nat} & nat,Prop& \textbf{None} & $2\times$forall& 2 \\
 \hline
 \lstinline?move : ? & Prop& \textbf{None}  & Hyp  & forall &1  \\
\hline
 \lstinline?move/ ? & Prop & \textbf{None} & andP & forall  & 1 \\
\hline
 \lstinline?rewrite? & \textbf{Prop} & $17\times$Prop & EL'' & equal, forall,equal&3\\
\hline 
\lstinline?case? &\textbf{-} &\textbf{-} &\textbf{-} &\textbf{-} &\textbf{-}\\
  \hline
 \lstinline?elim? &\textbf{nat} & \textbf{-} & \textbf{Hyp} & \textbf{forall}  & \textbf{1}\\
 \hline
  \end{tabular}
  
    \begin{tabular}{|l||l|l|l|l|l|}
\hline
 & \emph{arg-1 type} & \emph{rest arg types} & \emph{arg is hyp? }& \emph{top symbol} & \emph{n times used} \\
\hline
\hline
 \lstinline?move =>? & \textbf{nat} & \textbf{Prop}& \textbf{None} & \textbf{forall}& \textbf{1} \\
\hline
 \lstinline?move : ?&\textbf{-} &\textbf{-} &\textbf{-} &\textbf{-} &\textbf{-}\\
\hline
 \lstinline?move/ ? &\textbf{-} &\textbf{-} &\textbf{-} &\textbf{-} &\textbf{-}\\
\hline
 \lstinline?rewrite? & \textbf{Prop} & $6\times$Prop& EL'''  &$\mathbf{2\times}$\textbf{equal}&\textbf{2}\\
\hline 
\lstinline?case? &\textbf{-} &\textbf{-} &\textbf{-} &\textbf{-} &\textbf{-}\\
  \hline
 \lstinline?elim? &\textbf{nat} & \textbf{-}& \textbf{Hyp} & \textbf{equal}  & \textbf{1}\\
 \hline
  \end{tabular}}
  
 \caption{\footnotesize{\emph{\textbf{Tactic-level feature extraction table for \lstinline?sum_first_n?, \lstinline?sum_first_n_odd? and \lstinline?fact_prod? in SSReflect},
 showing correlation in bold: 15 out of 30 between \lstinline?sum_first_n? and \lstinline?sum_first_n_odd?; and 28 out of 30 between \lstinline?sum_first_n? 
 and \lstinline?fact_prod?. Where we use notation EL', EL'' and EL''', ML4PG gathers respectively the lemma names: (\lstinline?mul0n?, \lstinline?big_nat1?, \lstinline?muln0?,
 \lstinline?big_nat_recr?, \lstinline?mulnDr?, \lstinline?IH?, \lstinline?mulnD1?, \lstinline?addn2? and \lstinline?mulnC?), 
 (\lstinline?exp0n?, \lstinline?/index_iota?,  \lstinline?subn0?, \lstinline?big1_seq?, \lstinline?muln0?, \lstinline?in_nil?, \lstinline?big_mkcond?, \lstinline?addn1?, \lstinline?mulnDr?, \lstinline?muln1?, \lstinline?addn2?, \lstinline?big_nat_recr?, \lstinline?IH?,
   \lstinline?odd2n?, \lstinline?odd2n1?, \lstinline?addn0?, \lstinline?n1square? and \lstinline?n2square?) 
 and
 (\lstinline?big_nil?, \lstinline?factS?, \lstinline?big_add1?, \lstinline?IH?, \lstinline?big_add1?, \lstinline?big_nat_recr? and \lstinline?mulnC?); the lemmas and libraries can be found in \cite{HK12}. }}}
 \label{tab:mult_tactic}
\end{table}

\end{Exmp}

Finally,  ML4PG can extract the tree-level features, see Table~\ref{tab:mult_proof_tree}. Currently, it considers the proof flow using up to the depth 5 of the proof tree.


\begin{Exmp}
The tables for Lemmas \lstinline?sum_first_n?, \lstinline?sum_first_n_odd? and \lstinline?fact_prod? at the proof tree level are given in
Table~\ref{tab:mult_proof_tree}.
\begin{table}
\centering
\footnotesize{
\begin{tabular}{|l||l|l|l|l|l|l|l|l|l|l|l|}
\hline
 & \lstinline?move =>? & \lstinline?move :? & \lstinline?move/? & \lstinline?rewrite? & \lstinline?case? & \lstinline?elim?  &$\bullet$ & $\Box$ \\
 \hline
\hline
  \emph{td1} & - & - & - & - & - & nat & 12 & 0\\
 \hline
  \emph{td2} & nat, Prop& - & - & EL' & - &- &201 &1\\
 \hline
  \emph{td3} & - & - & - & EL''  & - & - & 30 & 1\\
 \hline 
  \end{tabular}

  \begin{tabular}{|l||l|l|l|l|l|l|l|l|l|l|l|}
\hline
 & \lstinline?move =>? & \lstinline?move :? & \lstinline?move/? & \lstinline?rewrite? & \lstinline?case? & \lstinline?elim? & $\bullet$ & $\Box$ \\
 \hline
\hline
  \emph{td1} & \textbf{-} & \textbf{-} & \textbf{-} & \textbf{-} & \textbf{-} & \textbf{nat} & \textbf{12} & \textbf{0}\\
 \hline
  \emph{td2} & \textbf{nat, Prop}& \textbf{-} & \textbf{-} & big\_nil  & \textbf{-}  &\textbf{-} &\textbf{201} &\textbf{1}\\
 \hline
  \emph{td3} & \textbf{-} & \textbf{-} & \textbf{-} &  EL'''  & \textbf{-} & \textbf{-} & \textbf{30} & \textbf{1}\\
 \hline 
  \end{tabular} }
 \caption{\footnotesize{\textbf{\emph{Proof tree level feature extraction table for SSReflect.}}  \emph{The rows, marked as \emph{td1--td5} represent tree levels.
 The values of columns  depend on the tactics, and for each tactic, a different parameter is tracked: 
 for tactics \lstinline?case? and \lstinline?elim? --- it is the type of argument; for tactic \lstinline?rewrite?  --- 
 whether the rewriting rule is a hypothesis, inductive hypothesis, or external lemma. Where we use notation EL', EL'' and EL''',
 ML4PG gathers respectively the lemma names:
 (\lstinline?mul0n?, \lstinline?big_nat1? and \lstinline?muln0?), 
 (\lstinline?big_nat_recr?, \lstinline?mulnDr?, \lstinline?IH?, \lstinline?mulnDl?, \lstinline?addn2? and \lstinline?mulnC?)
 and (\lstinline?factS?, \lstinline?big_add1?, \lstinline?IH?, \lstinline?big_add1?,  \lstinline?big_nat_recr? and \lstinline?mulnC?); 
 the lemmas and libraries can be found in \cite{HK12}.
In the $\bullet$ column ML4PG stores the branching factor of the proof tree. 
It encodes this information
globally using the following convention. The first number represents the
tree depth level, so, if we are at level $1$ the first number will be $1$, at level $2$
the first number will be $2$ and so on. Then for each one of the branches of the level
we store the number of its subbranches. 
The $\Box$ column  indicates the number of proof branches closed at the given level.
These tables show a fragment of proof tree level feature table for \lstinline?sum_first_n? and \lstinline?fact_prod?. We highlight in bold the
 corresponding features between 
\lstinline?sum_first_n? and \lstinline?fact_prod?
 (22 out of 24 features).}}}\label{tab:mult_proof_tree}
\end{table}
\end{Exmp}

The feature extraction procedures explained in this section run in the background of ML4PG during Coq compilation. Some of the features are obtained just by inspecting the names and numbers of the applied tactics.
In other cases, ML4PG needs to internally invoke
Coq compiler to obtain the features, for instance, when recording types of tactic arguments.  Thus, statistics related to the three proof levels is automatically gathered
during the proof development. 

Machine learning algorithms expect numerical feature vectors as inputs; therefore, ML4PG converts the features into numbers. 
As we explained in~\cite{KL12}, the concrete function that ML4PG uses for this purpose may vary, but the numeric conversion must be consistent. 
Dynamic calculation of the function that converts table features into numbers is implemented in ML4PG. In particular, we have defined
4 one-to-one functions $[[~.~]]_{Tactic}, [[~.~]]_{Type}, [[~.~]]_{Top\_symbol}$ and $[[~.~]]_{hyp\_or\_lemma}$ that assign respectively a numerical 
value to each tactic, type, top symbol and lemma appearing in a proof. The conversion provided by these functions is blind, assigning respectively unique consecutive integers
to tactics, types, top symbols and lemmas in order of their appearance in the library. If several elements appear in a cell, the value of that cell is the concatenation 
of the values of each element. 

\begin{Exmp}
The numerical table for Lemma \lstinline?sum_first_n_odd? at the goal level is given in
Table~\ref{tab:num_listnil}. This table is flattened into a vector to be given as input to machine learning algorithms. 
Namely, Table~\ref{tab:num_listnil} gives rise to the following feature vector:\\
$[3,1,-2,1,6,2,7,1,-444,126106,6,1,1517,4,-244444,112113105176,5,0,1,1,-24,0,5,1,7,1,$\\
$-44444444444,25484634152437143325432,6,0]$.

\begin{table}
 \centering
\footnotesize{
 \begin{tabular}{|l||l|l|l|l|l|l|}
 \hline
  & \emph{tactics} & \emph{N tactics} & \emph{arg type} & \emph{tactic arg is hypothesis?} & \emph{top symbol} & \emph{n subgoals} \\
 \hline
 \hline
 \emph{g1} & 3  & 1 & -2  & 1 & 6  & 2  \\
 \hline
   \emph{g2}& 7 & 1 &  -444 & 126106  & 6  & 1  \\
  \hline
   \emph{g3} & 1517  & 4 & -244444   &  112113105176 & 5  &0  \\
  \hline
 \emph{g4} & 1 & 1 & -24 &  0&  5 &1  \\	
  \hline
 \emph{g5} & 7 &  1 & -44444444444 & 25484634152437143325432  & 6 & 0 \\
  \hline 
   \end{tabular}}
  \caption{\footnotesize{\emph{\textbf{Numerical goal-level feature extraction table for \lstinline?sum_first_n_odd?.}}}}\label{tab:num_listnil}
  \end{table}

\end{Exmp}

%
Once the feature vectors are collected, ML4PG can data-mine the proofs using different machine learning
algorithms.

\section{Interactive Proof-Clustering in ML4PG}\label{sec:clustering}

ML4PG is designed to prove the concept: \emph{it is possible to interface higher-order proofs with machine learning engines, and do it interactively during the proof process}. 
Interaction with several machine learning engines
and algorithms is in the core of this process.  
This differs from  the experiments performed in the literature, see~\cite{Duncan02,KL12,KuhlweinLTUH12,UrbanSPV08}, where the data-mining
of proofs is performed separately from the user interface. 
In this section, we explain how ML4PG enables the user interaction with a range of 
machine learning engines and algorithms, and give some technical details of the ML4PG implementation.

The ML4PG user may or may not be familiar with machine learning. Either way, ML4PG 
 must offer him a number of simple but useful options to
 configure machine learning tools while staying within the Proof General environment. Therefore, ML4PG 
 takes the burden of connecting to the machine learning algorithms. 
 

 %

The first choice the user makes concerns the \emph{proof level}: proofs can be data-mined at the level of goals, tactics or proof trees, as explained in the
previous sections.
It is worth mentioning here that 
there are several choices of how to run this feature extraction. One option would be to extract features on demand --- that is, once the user chooses the proof level, ML4PG could re-run
Coq again to complete the feature extraction. The disadvantage of this  is that the proof engine will have to be re-run every time one uses ML4PG for data-mining.
We made a different choice: ML4PG extracts features in the background during the interactive proofs. 
It does the extraction at all three proof levels whenever the proof library is compiled.  Then, the choice of proof level in the menu just indicates which data set will be sent to the 
 machine learning algorithms. The advantage is that the time taken by data mining does not include the proof engine run. 
Our experiments show that the time involved in the feature extraction during the normal Coq compilation
  is unnoticeable, and does not  significantly slow down the proof development.

Now ML4PG is ready to communicate with \emph{machine learning interfaces}. 
ML4PG is built to be modular --- that is, when the feature extraction of Section~\ref{sec:alternative} is completed within the emacs environment,
the data is gathered in the format of hash tables. The first elements of these tables are the names of the lemmas, and the second elements are the feature
vectors encoded as lists of numbers (let us note that emacs is a Lisp environment; therefore, it is sensible to use lists to represent the feature vectors). 
However, every machine learning engine has its concrete format to represent feature vectors; therefore, it is necessary to define \emph{translators} to adapt
ML4PG's internal encoding of feature vectors to the concrete representation of the machine learning engine. 
We have defined translators for two different, but equally popular, machine learning interfaces --- MATLAB and Weka. 
ML4PG transforms the feature vectors
to a \emph{comma separated values} (csv) file in the case of MATLAB; and, to \emph{arff} files in the case of Weka. In principle, extending the list of machine learning
engines does not require any further modifications to the feature extraction algorithm, but just defining new \emph{translators}.
Notice the similarity with implementation of the proof level choice: again, once the features are extracted, the ML4PG engine is flexible to use them for all sorts of data mining tasks
and machine learning interfaces. 

Once the feature vectors are in a suitable format, ML4PG can invoke the machine learning engine. The ML4PG mechanism connecting to machine learning
interfaces is similar to the native mechanism of Proof General used to connect to ITPs. Namely, there is a synchronous communication between ML4PG and the
machine learning interfaces, which run on the background waiting for ML4PG calls.

 
 The next configuration option ML4PG offers is the choice of the particular \emph{pattern-recognition algorithm} available from the chosen machine learning interface. Again, this choice is made within the proof environment of Proof General.
There are several machine learning algorithms  available in MATLAB and Weka. We connected ML4PG only to  \emph{clustering algorithms}~\cite{Bishop} --- a family of
\emph{unsupervised learning methods}.  Unsupervised  learning is chosen when no user guidance or class tags are given to the algorithm in advance.

One could in principle envisage \emph{supervised machine learning applications} in proof pattern recognition, where the user labels every proof using some finite
tags, such as \lq\lq{}fundamental lemma\rq\rq{}, \lq\lq{}auxiliary lemma\rq\rq{}, \lq\lq{}proof experiment\rq\rq{}. And, on the basis of such labels and some number of training examples, the machine learning algorithm would be able to predict labels for any new proof.
Here, we do not assume existence of such labels. However, our modular approach to interfacing with Proof General implies that, if the labels are available, interfacing with supervised algorithms will not be hard for ML4PG. In fact, we envisage the feature extraction method to remain the same. Section~\ref{sec:rw} discusses related work using supervised learning in proof-pattern recognition.

In case of MATLAB, the algorithms included in ML4PG are the two most popular methods for clustering: K-means and Gaussian~\cite{XUWunsch05}.
If the user selects Weka as a machine learning engine; then, he can select among K-means, FarthestFirst and simple Expectation Maximisation.

 To improve the accuracy of the clustering algorithms, a technique called \emph{Principal Components Analysis} (PCA)~\cite{Joliffe86} is applied. This functionality 
 reduces the size of feature vectors but without much loss of information. The application of techniques like PCA, known
 in general as \emph{dimensionality reduction procedures}~\cite{XUWunsch05}, is recommended when dealing with feature vectors
 whose size is higher than $15$ --- as in our case.

%

Finally, the user can choose \emph{proof libraries} that he wants to access using ML4PG. Before using them, those libraries must be exported 
with the mechanism provided by ML4PG.  ML4PG extends the compilation procedure that Coq uses for imported libraries with the feature extraction algorithm
described above. Such a mechanism checks
that all the proofs of the library are finished, and generates a file which contains the list of the lemmas
of the library, and three files encoding respectively the feature vectors at the goal level, tactic level and proof tree level.
Subsequently, when the user chooses a library, ML4PG transforms the files to the internal encoding of feature vectors (implemented by Lisp lists) and attaches those
vectors to those obtained in the current development. 



By default, ML4PG clusters the current library, but the user can
add more libraries to perform clustering. The reason for not  using
all the available libraries is twofold. First of all, it is a question of \emph{performance}, since the time needed to
obtain clusters increases with the amount of libraries. The second reason is \emph{usability}, because if ML4PG uses all the
available libraries for clustering, it can obtain proof patterns from lemmas which belong to libraries unknown to the user, and this may or may not be convenient.

%
%

 
 

We now return to the examples introduced in Section~\ref{sec:levels} to illustrate how proof patterns are shown to the ML4PG user. 

\begin{Exmp}\label{example1}
We created a small library (70 lemmas) to help us with the initial tests of ML4PG: it contains some basic lemmas about natural numbers and lists, as well as our running examples of Tables \ref{tab:E12} and  \ref{tab:E5}. We also included efficient and inefficient proofs, 
and cases when similar lemmas were proven using different strategies, and different lemmas were proven using the same proof strategy.
In the rest of the paper, we will call the library \emph{\texttt{Initial}}.  
Figure~\ref{fig:clusters} shows the result of running ML4PG on this library, with the following settings:

\begin{itemize}
	\item statistics were taken using goal-level feature extraction;
	\item machine learning interface: Weka;
	\item machine learning algorithm: K-means.
\end{itemize}
ML4PG shows that all lemmas of Tables \ref{tab:E12} and  \ref{tab:E5} belong to
 the same group of proofs. It agrees with one possible interpretation of the content of these lemmas,  see \textbf{D.3} in Section~\ref{sec:levels}.
But there are other lemmas in that cluster; in particular, this cluster gathers ``fundamental'' lemmas about
various operations on natural numbers involving \verb"0" and operations on lists involving \verb"nil". 
 
 \begin{figure}
  \centering
  \includegraphics[scale=.4]{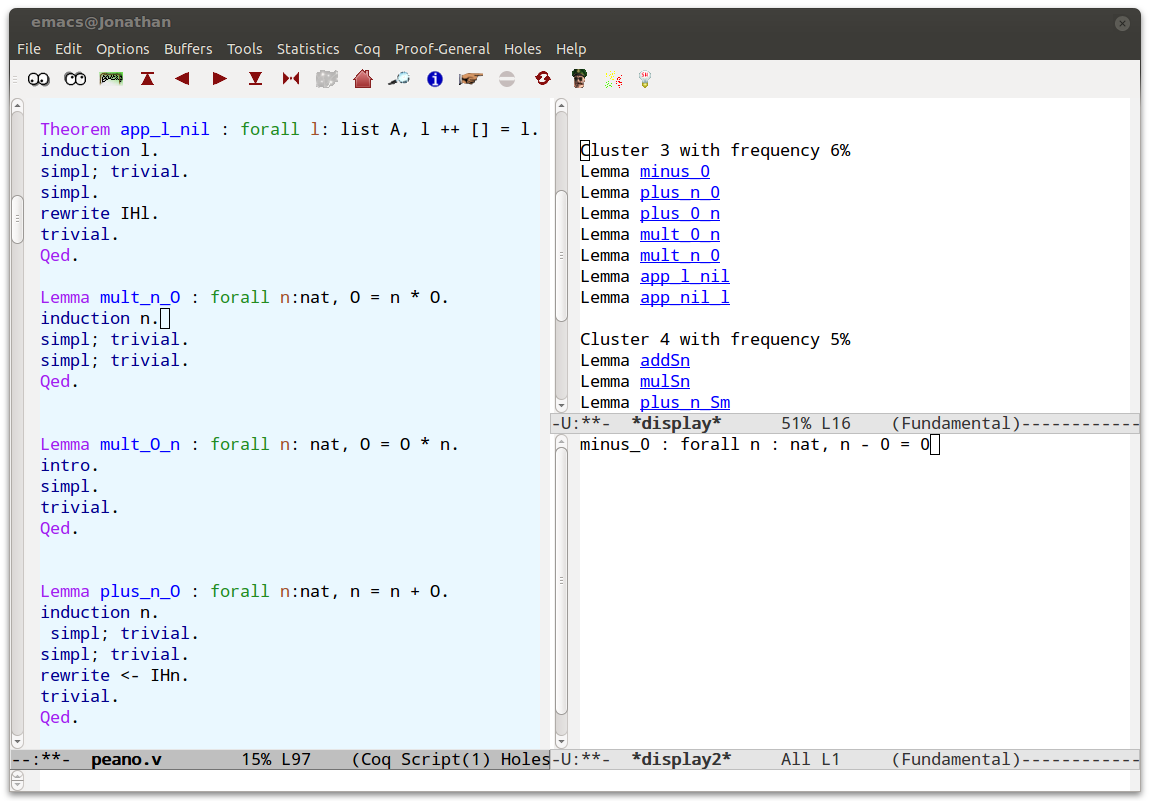}
  \caption{\footnotesize{\emph{\textbf{Clusters for the library \texttt{Initial}}. The Proof General window has been split into two windows
  positioned side by side: the left one keeps the current proof script, and the right one shows the clusters and their frequencies. If the user
  clicks on the name of a theorem showed in the right screen, such a window is split horizontally and a brief description of the selected theorem is
  shown.}}}\label{fig:clusters}
 \end{figure}
 
\end{Exmp}
   
 The example above shows one mode of working with ML4PG: that is, when a library is clustered irrespective of the current proof goal.
However, it may be useful to use this technology to aid the interactive proof development. In which case, we can cluster libraries relative to  a few 
initial proof steps for the current proof goal.  Note that ML4PG graphical interface offers two menu buttons for these two options --
the two right-most  buttons of Figure~\ref{fig:clusters}. The next example illustrates this.
 
 \begin{Exmp}\label{example2}
  On the left side of Figure~\ref{fig:similarities4}, an incomplete development of Lemma \lstinline?sum_first_n? is shown. 
  Using the \lstinline?bigop? and \lstinline?binomial? libraries of SSReflect (205 lemmas), ML4PG can obtain proof patterns similar to the 
  lemma that we are proving, see right side of Figure~\ref{fig:similarities4}. The ML4PG settings in this case were:
  
  \begin{itemize}
	\item statistics was taken using goal-level feature extraction;
	\item machine learning interface: MATLAB;
	\item machine learning algorithm: K-means.
\end{itemize}
  
 Among the suggestions provided by ML4PG, we find the Lemma \lstinline?fact_prod?.
 Note that,  as we have seen in Section~\ref{sec:alternative}, there is a high correlation between their feature vectors. 
  Other lemmas ML4PG discovered are related to series of natural numbers (including properties about big sums and big products). Lemmas like \lstinline?sum_first_n_odd?,
  where there is a restriction on the elements of the series,  belong to a different cluster since the correlation with lemmas like  \lstinline?sum_first_n? 
  and \lstinline?fact_prod? is low. 
  
  %
  
  \begin{figure}
  \centering
  \includegraphics[scale=.4]{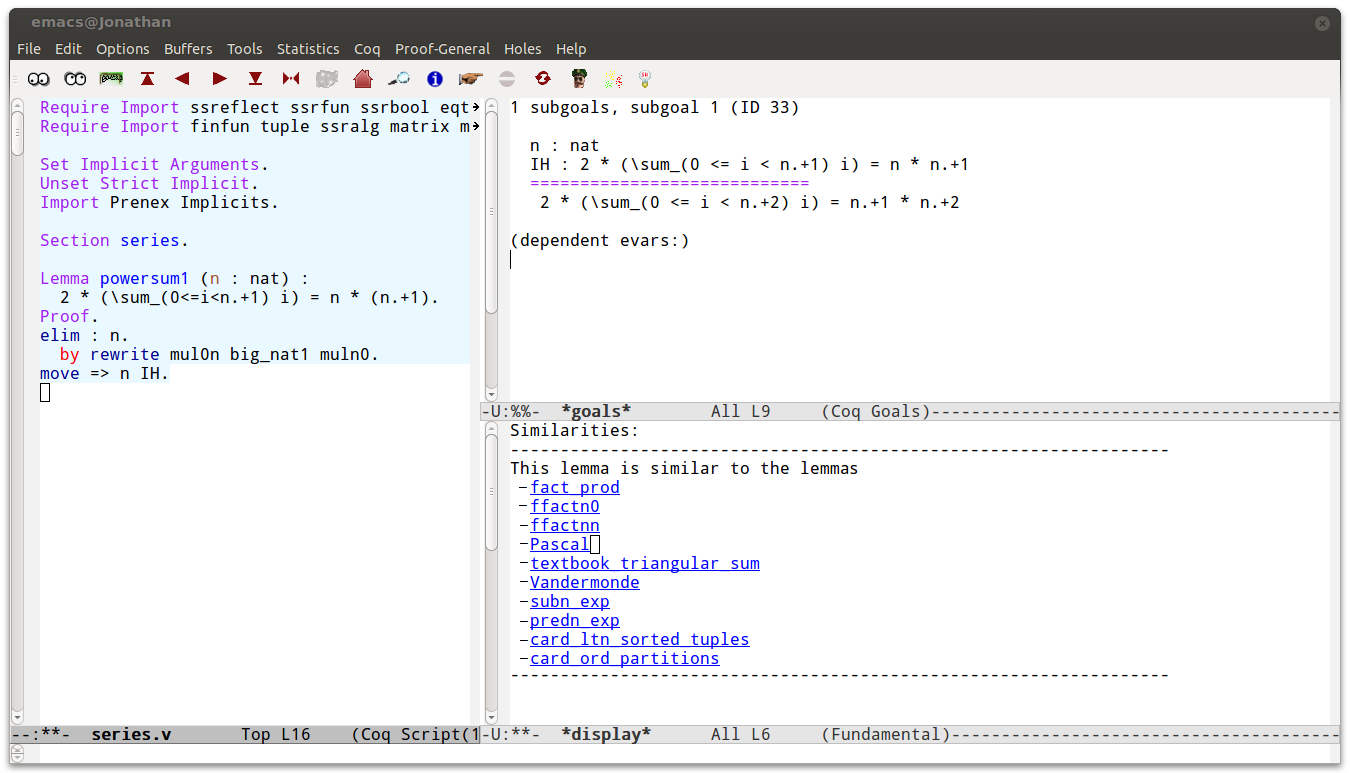}
  \caption{\footnotesize{\emph{\textbf{Similarities of Lemma \lstinline?sum_first_n?.} On the left side, the development about the Lemma \lstinline?sum_first_n?. On the top part of the right side, the current goal. On
  the bottom part of the right side, several suggestions provided by ML4PG. If the user clicks on the name of one of the suggested lemmas, 
  a brief description about it is shown.}}}\label{fig:similarities4}
\end{figure}

  \end{Exmp}

We have shown flexibility, modularity and interactivity of ML4PG in interfacing with machine learning environments. 
These features come for free  with the light version of \emph{\lq\lq{}backward interfacing\rq\rq{}} that ML4PG implements:
that is, it does not translate the outputs of the clustering algorithms back into the Coq language. Its only form of backward interfacing is 
conversion of clustered feature vectors back into lemma names --- the output shown in  Figures~\ref{fig:clusters} and \ref{fig:similarities4}.

Generally,  interfacing the ITPs with external tools (e.g. ATPs) is a challenging task,
 see~\cite{MP09,HT98,GAPCOQ,Laurent1}. A special concern is the 
translation of the output produced by the external tools into the ITP. 
This is due to the fact that unsound translation can introduce inconsistencies in the system; see e.g.~\cite{TACAS13}.
In case of interfacing with machine learning, the external tool is even more alien to ITP\rq{}s syntax than ATPs.  
Light backward interfacing implemented in ML4PG may well be the optimal solution to the problem. 

\section{Handling Proof Statistics in ML4PG}\label{sec:session}

The previous sections highlighted two features of ML4PG --- light backward interfacing and interactive handling of machine learning interfaces. 
To handle these features gracefully, ML4PG must offer the user a convenient environment
for processing and analysing the \emph{results} obtained by machine learning algorithms.


Clustering techniques divide data into $n$ groups of similar objects (called clusters), where the value of $n$ is a ``learning'' parameter provided by the user together with other inputs to the clustering algorithm.
Increasing the value of $n$ means that the algorithm will try to separate objects into more classes, and, as a consequence, each cluster will contain fewer examples with higher correlation. 
 The frequencies of clusters can serve for analysis of their reliability. 
Results of one run of a clustering algorithm may differ from another, even on the same data set.
This is due to the fact that clustering algorithms randomly choose examples to start from, and 
form clusters relative to those examples. However, it may happen that certain clusters
are found repeatedly --- and frequently --- in different runs; then, we can use these frequencies to determine the reliable clusters.

ML4PG's tools handling statistical results include  a set of programs written in 
MATLAB and Weka,  that post-process outputs of the clustering algorithms.
For each clustering algorithm the user invokes, ML4PG generates one corresponding program handling the output statistics. 
These various programs always have three arguments: a file and two natural numbers representing the number of clusters and frequency threshold. We explain these settings in this section. 


 Various numbers of clusters can be useful for interactive proof data-mining: this may depend on the 
size of the data set, and on existing similarities between the proofs. 
We want ML4PG to accommodate such choices. 
In general, small values of $n$ are useful when searching for general proof patterns which can later be refined by increasing the value of $n$.
However, extreme values are to be avoided: small values of $n$ can produce meaningless proof clusters for big proof libraries; whereas trivial clusters with just one proof
may be found for big values of $n$. Very often in machine learning, the optimal number of clusters is determined experimentally, but we cannot afford this in ML4PG setting, as we
 assume the user focuses mainly on the Coq proofs. 

In the machine learning literature, there exists a number of heuristics to determine this optimal number of clusters,~\cite{XUWunsch05}.
We used them as an inspiration to formulate our own algorithm for ML4PG, tailored to the interactive proofs. 
At any given stage of the interactive proof, it takes into consideration the size of the proof library 
and an auxiliary parameter we introduce here, called \emph{granularity}. This parameter is used to 
calculate the optimal number of proof clusters, using the formulas of Table~\ref{tab:granularity}. As a result, the user does not provide the value of $n$, but
just decides on \emph{granularity} in ML4PG menu, by selecting a value between $1$ and $5$, where $1$ stands for a low 
granularity (producing big and general clusters) and $5$ stands for a high granularity (producing small and precise clusters).

\begin{table}
  \centering
  \begin{tabular}{cc}
   \footnotesize{
\begin{tabular}{|c | c|}
\hline
Granularity~~ & ~~Number of clusters~~  \\ 
\hline
\hline
1  &  $\lfloor l/10 \rfloor$\\
2  & $\lfloor l/9 \rfloor$          \\
3 & $\lfloor l/8 \rfloor$       \\ 
4 & $\lfloor l/7 \rfloor$      \\
5 & $\lfloor l/6 \rfloor$   \\
\hline
\end{tabular}}
&
\footnotesize{\begin{tabular}{|c | c|}
\hline
Frequency parameter~~ & ~~ Frequency Threshold~~  \\ 
\hline \hline
1  &  $5\%$\\
2  & $15\%$          \\
3 & $30\%$       \\ 
\hline
\end{tabular}
}

  \end{tabular}

\caption{{\footnotesize \emph{\textbf{ML4PG formulas computing clustering parameters.} \textbf{Left:} the formula computing the number of clusters given the 
granularity value, where $l$ is the number of lemmas in the library.
\textbf{Right:} the formula computing frequency thresholds given a frequency parameter.}}}\label{tab:granularity} 
\end{table}


\begin{Exmp}\label{ex:gran}
Consider Example~\ref{example1}: there, the default granularity was $3$, and the cluster contained all lemmas from Tables~\ref{tab:E12} and~\ref{tab:E5}.
Increasing the granularity, ML4PG discovers only Lemmas \lstinline?app_l_nil? and \lstinline?mult_n_0? (see \textbf{D.1} from Section \ref{sec:levels}), as well as similar proofs for \lstinline?plus_n_0? and \lstinline?minus_n_0?. 
All of them use induction, and prove similar base cases. Note that in Section  \ref{sec:levels} we conjectured this separation of  examples of Tables~\ref{tab:E12} and~\ref{tab:E5} into two clusters  as a desirable feature. 
\end{Exmp}

\begin{Exmp}
 In Example~\ref{example2}, ML4PG used the default granularity value of $3$, to obtain ten suggestions related to the Lemma \lstinline?sum_first_n?. 
 If the ML4PG user increases such granularity value to $5$, he obtains only one suggestion, see Figure~\ref{fig:similarities3}: the Lemma \lstinline?fact_prod?.
 Inspecting the proof of Lemma \lstinline?fact_prod? can give an insight into how to finish the proof for  \lstinline?sum_first_n?. We notice that we can apply the Lemma \lstinline?big_nat_recr? to our current goal and, subsequently use the
 inductive hypothesis. The rest of the proof is based on rewriting rules of natural numbers. 
 
 \begin{figure}
  \centering
  \includegraphics[scale=.4]{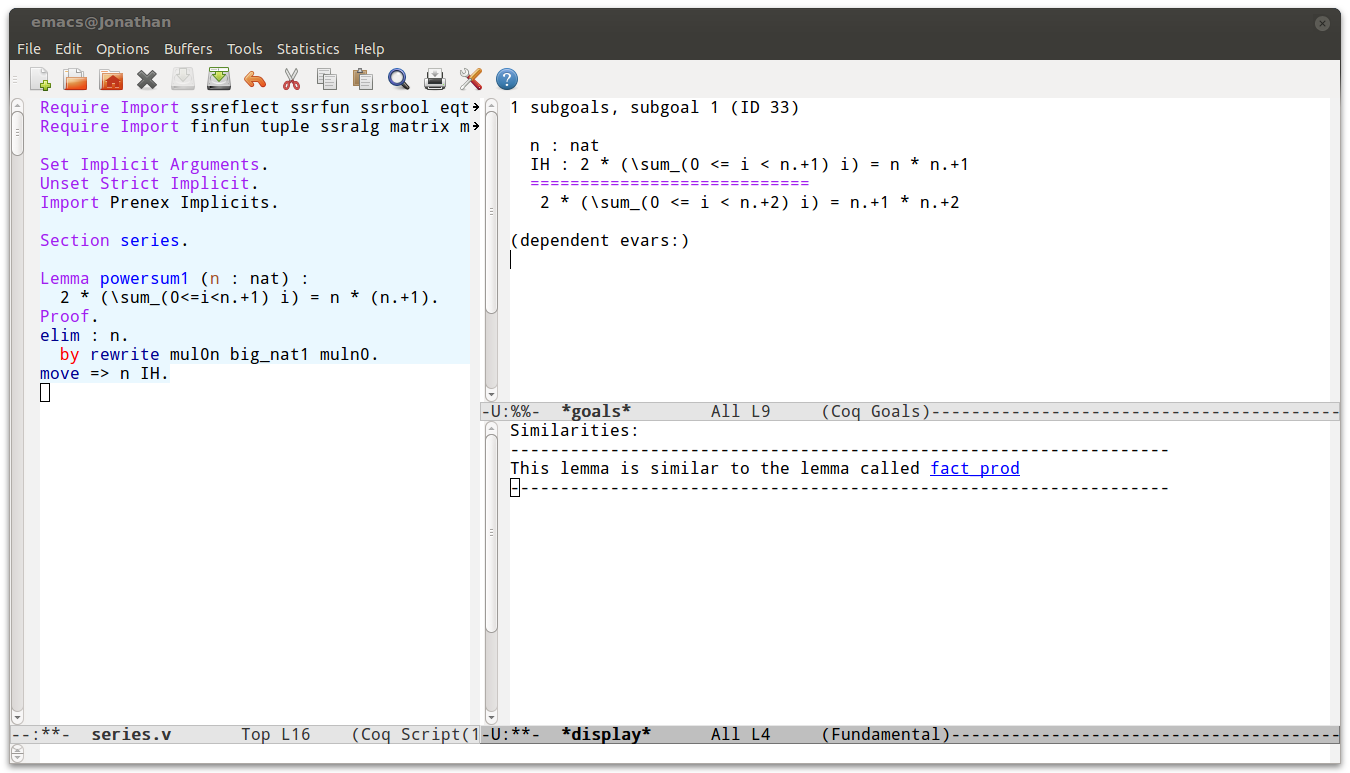}
  \caption{\footnotesize{\emph{\textbf{Suggestion about the Lemma \lstinline?sum_first_n? using  the granularity value $5$.}}}}\label{fig:similarities3}
 \end{figure}
 
\end{Exmp} 
 
 As implied by the above examples, the configuration of the granularity parameter can be approached in two different ways: \emph{top-down} and \emph{bottom-up}.
 The top-down approach suggests first 
 using a small value for the granularity to obtain a general proof pattern, and then refine that pattern increasing the granularity value. 
 On the contrary, in the bottom-up approach a high value for the granularity is used to see what the most similar lemmas are and then decrease the
 granularity value to see more general --- and potentially less trivial --- patterns. 
 
Finally, the third parameter ML4PG uses to analyse clustering outputs is the \emph{frequency threshold}. For this purpose, ML4PG actually uses 
 double criteria: the \emph{proximity} and  \emph{frequency} of the cluster. The
clustering algorithm output contains not only clusters but also a proximity value ---  a measure of how close each object in one cluster is with respect to objects in other clusters.  
This measure ranges from $+1$, indicating points that are very distant from other clusters, through $0$,
indicating points that are not distinctly in one cluster or another, to $-1$, indicating points that are probably assigned to the 
wrong cluster. We have fixed $0.5$ as an accuracy threshold, and all the clusters whose measure is under such value are ignored by ML4PG. This criterion is
fixed, and the user interface does not give access to it. 
However, the second criterion, the frequency parameter, is customizable within the interface. 

Our experience shows that analysis of frequencies may give two opposite effects. 
\begin{itemize}
\item[*] On the one hand, high frequencies suggest that the proofs  found in clusters 
have a high correlation, and that is a desirable property. 
\item[**] On the other hand, proofs with too high correlation may be too trivial for providing interesting proof hints.
Therefore, it is sometimes useful to look for proof clusters with lower frequencies --- as they may potentially contain those non-trivial analogies.
\end{itemize}

\begin{Exmp}\label{example3}
Illustrating this,
in our running example, the four proofs from Tables~\ref{tab:E12} and~\ref{tab:E5} were initially found only in $6\%$ of runs (low frequency),
see Figure~\ref{fig:clusters}; whereas there were other clusters with high frequencies that contained trivially similar lemmas (for instance, a cluster contains all the 
lemmas of the shape $x + 0 = x$ where $x$ is a term --- the proofs of all these lemmas are the same). 
\end{Exmp}


To gather sufficient statistical data from proofs,
ML4PG runs  the chosen clustering algorithm $200$ times  at every call of clustering, and collects the frequencies of each cluster. After discarding those with low proximity, it 
calculates the frequency of the significant patterns. Once frequencies are calculated, ML4PG applies the following methodology.
 As item * suggests, one purpose of the frequencies is to serve as thresholds: if the number of times the cluster occurs falls below the pre-set threshold, 
the corresponding proofs will not be displayed to the user. 
On the other hand, as item ** suggests, the acceptable \emph{frequency threshold} values may differ from proof to proof, and may depend on the purpose of proof pattern recognition.
For this, ML4PG allows the user to vary the threshold values. At the moment, we implemented three choices: frequency parameters 1--3 as shown in Table \ref{tab:granularity}. This particular range of
 thresholds comes from our experience with several Coq and SSReflect libraries. However, in line with our general modular approach to ML4PG design, we 
assume a wider range can be implemented, if desired. Our current choice is to keep the ML4PG interface simple and minimalistic.



\begin{Exmp}
 Table~\ref{tab:res} shows the results for different choices of algorithms and parameters for Example~\ref{example2}, we 
 highlight the result presented in Figure~\ref{fig:similarities4}.

\end{Exmp}

 \begin{table}
 {\scriptsize 
  \centering
  \begin{tabular}{|c|c|c|c|c|c|c|c|c|c|}
   \hline
    & $g=1$&$g=1$ &$g=1$ & $g=2$&$g=2$ &$g=2$ & $g=3$&$g=3$ &$g=3$  \\
    & $f=5\%$&$f=15\%$ &$f=30\%$ & $f=5\%$&$f=15\%$ &$f=30\%$ & $f=5\%$&$f=15\%$ &$f=30\%$  \\
    & $n=20$& $n=20$ & $n=20$ & $n=22$& $n=22$ & $n=22$ & $n=25$& $n=25$ & $n=25$  \\
   \hline
   Gaussian & 99 &0  &0  &76  & 76 &0  &30  &30  & 30   \\
   \hline
   K-means (MATLAB) & 30 & 30  & 30 & 21  &  21 & 21 & 10 & \textbf{10} &10   \\
   \hline
    K-means (Weka) & 26 & 26  & 0 & 22  &  22 & 22 & 20 & 20  &20   \\
   \hline
   Expectation Maximisation &80  & 0 &0  &72  &0  &0  &64  &64  &0   \\
   \hline
   FarthestFirst &0  &0  & 0 & 81 & 0 &0  &73  &0  &0  \\
   \hline
  \end{tabular} 
    \begin{tabular}{|c|c|c|c|c|c|c|}
   \hline
     & $g=4$&$g=4$ &$g=4$ & $g=5$&$g=5$ &$g=5$ \\
    & $f=5\%$&$f=15\%$ &$f=30\%$ & $f=5\%$&$f=15\%$ &$f=30\%$  \\
    & $n=29$& $n=29$ & $n=29$ & $n=34$& $n=34$ & $n=34$ \\
   \hline
   Gaussian  & 8 &8  &8  &1  &1  &1  \\
   \hline
   K-means (MATLAB) &  7 & 7  &7  & 1 &\textbf{1}  &1  \\
   \hline
   K-means (Weka) &  20 & 20  &20  & 11 & 11  & 11  \\
   \hline
   Expectation Maximisation  & 20 &20  &20  &3  &3  &3  \\
   \hline
   FarthestFirst  &55  &55  &0  &20  &20  & 20 \\
   \hline
  \end{tabular}}   
  \caption{\footnotesize{\emph{\textbf{A series of clustering experiments for Example~\ref{example2}}. The rows of the table indicate the algorithm used by ML4PG, and the columns show 
  the values of the parameters: granularity ($g$), frequency thresholds ($f$) and the number of clusters $n$ (computed from the granularity parameter). We record here only the clusters
	containing Lemma \lstinline?fact_prod? --- the most ``useful'' proof hint in the context. Note that such clusters are found irrespective of the chosen algorithm. The size of the data set is 205 lemmas.
	The results in bold provide the settings to obtain the result presented in Figures~\ref{fig:similarities4} and~\ref{fig:similarities3}}}} \label{tab:res}
 \end{table}

Our next example shows an interesting interplay between the effects of varying granularity, frequency, and proof level in the process of proof data-mining.

\begin{figure}[h]
  \centering
\begin{tabular}{ccc}
 
  \includegraphics[scale=.4]{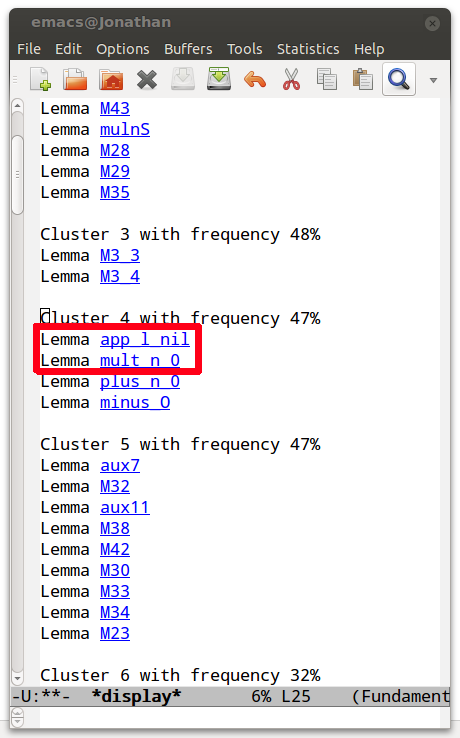}
&  
 \includegraphics[scale=.4]{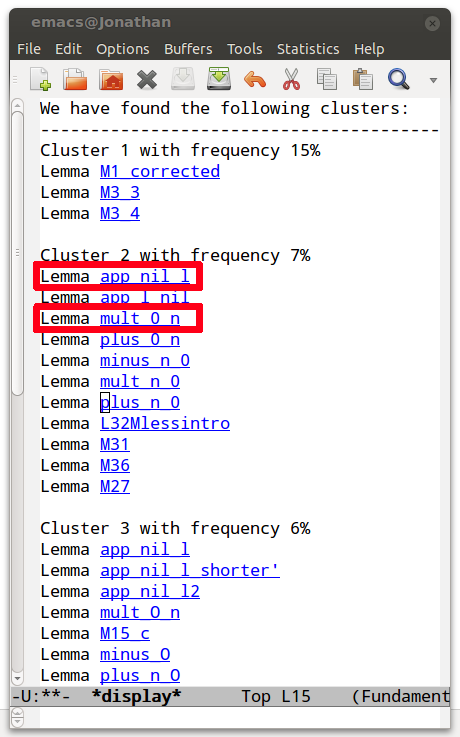}
&  
 \includegraphics[scale=.4]{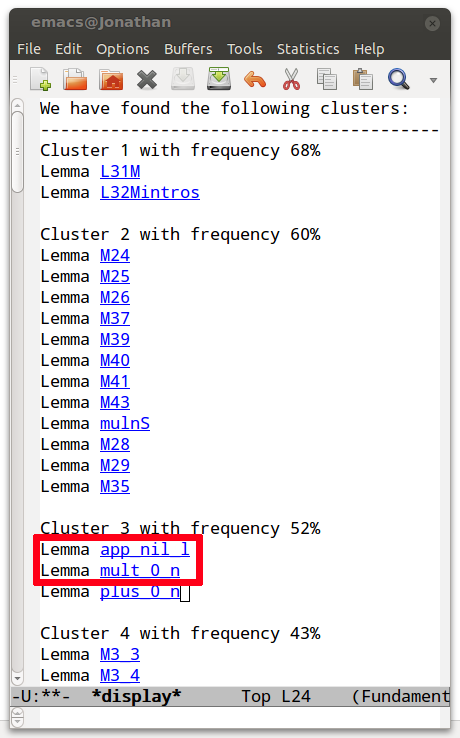}

\end{tabular}
  \caption{\footnotesize{\emph{\textbf{Effects of increasing granularity of clustering on the levels of goals and tactics for the library \emph{\texttt{Initial}}.} \textbf{Left:} 
	compare to Figure \ref{fig:clusters}, smaller clusters are formed when $5$ is chosen  as the granularity  value at the goal level; frequencies increase.
	As we conjectured in Item \textbf{D.1} of Example~\ref{ex:example0}, Lemmas \texttt{app\_l\_nil} and \texttt{mult\_n\_0} belong to the same cluster (see the lemmas highlighted
	with a circle in the screenshot). 
	\textbf{Centre:} Tactic-level clusters formed with granularity parameter  $2$; the frequencies are relatively low. 
	As we conjectured in Item \textbf{D.2} of Example~\ref{ex:example0}, Lemmas \texttt{app\_nil\_l} and \texttt{mult\_0\_n} belong to the same cluster 
	when considering the tactic-level (see the lemmas highlighted 	with a circle in the screenshot). 
\textbf{Right:} Tactic-level clusters formed with granularity parameter $5$, the clusters become smaller, and the frequencies increase. Even increasing the 
granularity, Lemmas \texttt{app\_nil\_l} and \texttt{mult\_0\_n} are still in the same cluster.}}}\label{fig:new}
\end{figure}

\begin{Exmp}
 In Example~\ref{example1}, ML4PG shows the clusters for our four running examples from the library \texttt{Initial}, when using the default frequency value of $1$. 
As Example~\ref{example3} showed, when increasing the frequencies parameter, such a cluster would fall below the threshold (compare with  Figure~\ref{fig:clusters}, where the
 frequency of this cluster was $6\%$). At the same time, when increasing the granularity parameter to $5$, our four proofs will be split into two smaller clusters, each having higher frequencies. 
 Notably, inductive proofs (see \textbf{D.1} and Table \ref{tab:E12}  from Section \ref{sec:levels}  and Example \ref{ex:gran}) are separated from those by simplification (see \textbf{D.2} and Table \ref{tab:E5}). 
The cluster containing only lemmas from Table \ref{tab:E12} has a frequency of $47\%$ (see the left screenshot of Figure~\ref{fig:new}).
The proofs from Table~\ref{tab:E5} also form a smaller cluster, but with frequency of $7\%$.
 Therefore, both clusters are shown if the frequency parameter is $1$, but we also have an option of choosing a higher frequency ($15 \%$ or $30 \%$) to 
discard the second, less significant, cluster. 

 \end{Exmp}

 This is a typical situation, small values of the granularity parameter usually produce big clusters with small frequencies.
When the granularity value is increased, the big clusters are split into smaller ones with high frequencies. Note the interactive nature of this proof pattern recognition process.

\begin{Exmp} A similar effect of increasing granularity parameter and increasing frequencies for the tactic-level proof features is shown in Figure \ref{fig:new}.
The figure also demonstrates that data-mining the same library using goal-level features and tactic-level features can bring different results. 
Interestingly, with increase of granularity, the goal-level clustering focuses on the examples related to lemmas in Table \ref{tab:E12}, whereas the tactic-level clustering focuses
on examples related to lemmas of Table \ref{tab:E5}; as we conjectured in items \textbf{D.1} and \textbf{D.2} of Section \ref{sec:levels}.  
\end{Exmp}

\begin{figure}
  \centering
  \includegraphics[scale=.5]{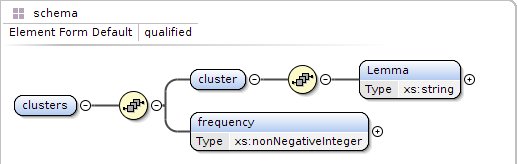}
  \caption{\footnotesize{\emph{\textbf{XML schema for handling results produced by different machine learning engines.} An XML file following
  this schema will consists of a main node called clusters, which has as child a sequence of pairs. The first element of the 
  pair is a node called cluster whose child is the sequence of lemmas  belonging to the cluster. The second element of the pair
  is the frequency of the cluster.}}}\label{fig:schema}
\end{figure}

We finish this section with a discussion of the role of this statistical analysis in our approach to the \emph{light backward interfacing}.
ML4PG handles  the results obtained with MATLAB and Weka in a uniform way: for this purpose, we devised an XML format, see Figure~\ref{fig:schema}. 
Using this approach, ML4PG can deal with the output generated by any system which follows this XML standard using just one program which transforms the XML
files into a suitable format for the user. As a consequence, ML4PG can be easily extended with new engines and machine learning algorithms. 

The XML files returned by the machine learning engines are processed in two different ways depending on the mode of using ML4PG: that is, general clustering  (as 
illustrated in Example \ref{example1}) or goal-dependent clustering (as shown in Example \ref{example2}).
In both cases, the XML file is converted to a list of pairs where the first element of the pair contains the lemma names and the
second element the frequency of each cluster. In the general clustering case, such a list is processed to be shown as in Figure~\ref{fig:clusters}.
For the goal-dependent clustering, ML4PG searches for those pairs of the list where the current proof is included.
If the current proof belongs to several clusters, then ML4PG takes the one with the
highest frequency and displays it as shown in Figure~\ref{fig:similarities4}.

\section{Integrating machine learning with theorem proving: related work}\label{sec:rw}

In this section, we present an overview of the integration of machine learning techniques into
automated and interactive proofs. There are two machine learning styles that can be useful in
this context: \emph{symbolic} and \emph{statistical}.

\emph{Symbolic machine learning} methods can formulate auxiliary lemmas or proof strategies from background 
knowledge. As we have explained in the introduction, Proof General is a general-purpose 
interface for a range of higher-order theorem provers, and this is probably the reason 
why this interface has been used in different works to integrate machine learning techniques. 
This is the case of IsaPlanner~\cite{DixonF03}, a generic framework for proof planning that integrates
techniques like rippling~\cite{BB05} in the interactive theorem prover Isabelle. 
Also using Isabelle as a prover and Proof General as interface, PGtips~\cite{PGtips} is a tactic recommender 
system based on data-mining techniques~\cite{Duncan02}. Another advisor implemented in Proof General, but 
in this case for the Mizar system, was incorporated into MizarMode~\cite{MizarMode} to suggest similar
results which could be useful in the current proof. 

Inductive provers like ACL2~\cite{KMM00} and Hip~\cite{Rosen12} also include symbolic machine learning 
techniques. In particular, ACL2 uses these methods for proving termination of programs written in Lisp~\cite{CCGACL2},
and Hip implements a theory discovery mechanism~\cite{CJRS13} for automatically derive and prove properties about 
functional programs implemented in Haskell.

The main drawback of the symbolic methods is their scope. These techniques are often tailored for certain 
fragments of first order language or a certain library, or a certain proof shape; therefore, they do not
scale properly to deal with big libraries. On the other hand, statistical machine learning methods scale to big 
libraries without problems, but they have very weak power for generalisation. 

\emph{Statistical machine learning} methods can discover data regularities based on numeric proof features. 
ML4PG belongs to this category of methods. Among other successful statistical tools is the method of 
statistical proof-premise selection~\cite{Kuehlwein2012,TsivtsivadzeUGH11,KuhlweinLTUH12,Mash}. Applied in several ATPs, it provides 
statistical ratings of existing lemmas; and this makes automated rewriting more efficient.

This technique has also been used to integrate ITPs, ATPs and machine learning.
The workflow of tools like Sledgehammer to prove a theorem consists of the following steps: (1) translation of the statement of the theorem (from Isabelle, HOL or Mizar format)
to a first order format suitable for ATPs; (2) selection of the lemmas (or premises) that could be relevant to prove the theorem; (3) invocation of several ATPs to prove the result; and (4)
if an ATP is successful in the proof, reconstruction of the proof in the ITP from the output generated by the ATP. An important issue in this procedure is the 
premise selection mechanism, especially when dealing with big libraries, since proofs of some results can be infeasible for the ATPs if they receive too many premises.  

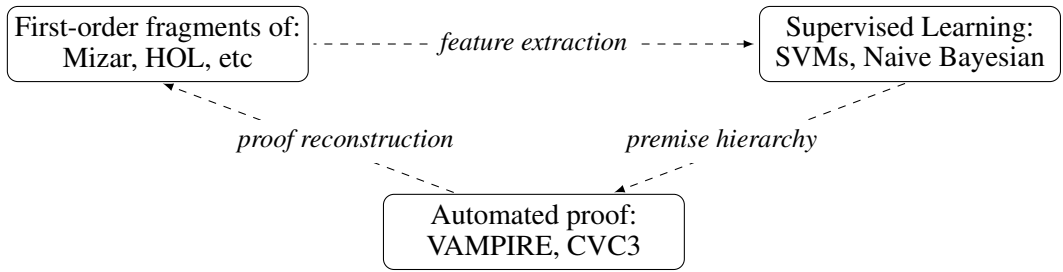
\begin{figure}
\begin{center}
 \begin{tikzpicture}
\draw[fill=white,rounded corners] (-2,.5) rectangle (2,-.5); 
\draw (0,0.2) node{First-order fragments of:}; 
\draw (0,-0.2) node{Mizar, HOL, etc}; 

\draw[fill=white,rounded corners] (8,.5) rectangle (12,-.5); 
\draw (10,.2) node{Supervised Learning:};
\draw (10,-.2) node{SVMs, Naive Bayesian};

\draw[latex-,shorten <=2pt,shorten >=2pt,dashed] (8,0) -- (2,0); 
\draw (5,0.3) node[anchor=north,fill=white]{\emph{\small{feature extraction}}};

\draw[fill=white,rounded corners] (3,-2) rectangle (7,-3); 
\draw (5,-2.3) node{Automated proof:} ;
\draw (5,-2.7) node{VAMPIRE, CVC3} ;

\draw[dashed,shorten <=2pt,shorten >=2pt,-latex] (4,-2) -- (0,-.5);
\draw (2.5,-1) node[anchor=north,fill=white]{\emph{\small{proof reconstruction}}};

\draw[dashed,shorten <=2pt,shorten >=2pt,-latex] (10,-.5) -- (6,-2); 
\draw (7.5,-1) node[anchor=north,fill=white]{\emph{\small{premise hierarchy}}};
 
\end{tikzpicture}
\end{center}
\caption{\emph{\footnotesize{\textbf{Machine learning techniques for premise selection.}}}}\label{fig:urban}

\end{figure}

Statistical machine learning methods are used to tackle this problem in~\cite{KuhlweinLTUH12,Mash}. In particular, a classifier is constructed for every 
lemma of the library. Given two lemmas $A$ and $B$, if $B$ was used in the proof of $A$, a feature vector $|A|$  is extracted 
from the statement of $A$, and is sent as a positive example to the classifier $<B>$ constructed for $B$; otherwise, $|A|$ is
used as a negative example to train $<B>$.
Note that, $|A|$  captures statistics of $A$'s syntactic form relative to
every symbol in the library; and the resulting feature vector is a sparse (including up to $10^6$ features).
 After such training, the classifier $<B>$ can predict whether a new lemma $C$ requires 
the lemma $B$ in its proof, by testing $<B>$ with the input vector $|C|$. On the basis of such predictions for all lemmas in the library,  this tool constructs a hierarchy of the premises that are most likely to be used in the proof
of $C$. Figure~\ref{fig:urban} illustrates this approach. 

Table~\ref{tab:diff} summarises the main differences between premise-selection tools and ML4PG. It is important to notice that the two methods have different approaches 
to handling large-size data.
Premise-selection tools rely on advanced \emph{``sparse''} machine-learning algorithms to process the growing feature vectors. ML4PG achieves scaling at the stage 
of feature extraction, by using 
the proof trace method to produce compact feature vectors.
As a result, ML4PG also works well with libraries of small size (and hence can interact with the user at any stage of the proof), 
whereas sparse methods need proof libraries of big size to perform well.  

\begin{table}
 \centering
 \begin{tabular}{|c||c|c|}
  \hline
  & premise-selection tools &  ML4PG \\
  \hline
  \hline
  Aim & increase number of goals  & assist the user providing\\
     & automatically proved by ATPs  & proof families as proof hints\\
  \hline
  
  Features & extracted from first-order formulas & extracted from higher-order  \\
   & & formulas and proofs\\
  \hline
    
  Feature vectors & sparse & dense\\
  representation & (up to $10^6$ features)& ($\approx 30$ features)\\
  \hline
  
  Machine learning  & supervised learning & unsupervised learning\\
  methods & (SVMs, Naive Bayesian, \ldots) & (K-means, Gaussian, \ldots) \\
  \hline
  
	 Scaling for big   & on machine-learning level: & on the interface level:\\
  libraries & \emph{sparse} algorithms & \emph{proof-trace} method \\
  \hline
  
  \hline
 \end{tabular}

 \caption{\emph{\footnotesize{\textbf{Differences between premise-selection tools and ML4PG.}}}}\label{tab:diff}

\end{table}

\section{Conclusions and Further work}\label{sec:conclusions}

In this paper, we have presented a Proof General extension, called ML4PG, to interface ITPs and machine learning 
engines. Our main goal was to prove that it is possible to interface higher-order \emph{interactive} theorem proving with statistical machine learning;
and the resulting tool can provide fast and non-trivial proof hints during the proof development. The technical highlights of ML4PG are:
\begin{itemize}  
\item the \emph{proof trace method} is a flexible, extendable technique that 
gathers statistics from proofs on the basis of the relative transformations of simple parameters within several proof steps; and, 

\item the \emph{light backward interfacing} implemented in ML4PG automates the Proof General interaction with machine learning engines. It helps to analyse and interpret the output of
machine learning algorithms; however, it avoids full translation of the statistical outputs into the prover\rq{}s language. 
\end{itemize}

The ML4PG approach has several benefits. First of all, it does not assume any knowledge of machine learning interfaces from the user; and automates initial statistical experiments (determining the number of clusters, calculating frequencies) that otherwise would have been performed by hand. 
The choices for various measures of cluster granularity and frequency can be easily extended in the future.
Moreover, it is a modular tool which allows the user to make choices regarding approach to levels of proofs and particular statistical
algorithms. By design, it allows further extensions to different machine learning environments, modes of supervised/unsupervised learning, and various learning algorithms within those modes.
In addition, it is tolerant to mixing and matching different proof libraries, different notations and proof styles used 
across several developments. 

Comparing across different proof levels and different styles of proofs, our experiments show that data-mining the goal-level features 
shows more interesting clusters compared to the other two feature extraction methods. 
We plan to improve  the other two feature extraction methods in the future.
Proofs in SSReflect yield more consistent classification results compared to the plain Coq style.
This is due to a stricter proof discipline in SSReflect, which allows ML4PG to detect more significant proof patterns. 

The feature extraction method implemented in ML4PG can be improved in two different ways. First of all, the proof trace
method considers just the first five proof steps, losing some information. As the machine learning algorithms integrated in 
ML4PG require a fixed number of features whereas Coq proofs may have varied length, we can study big proofs considering either small patches of 
proofs (by partially reusing the proof trace method), or proofs as a whole (in this case, a sparse representation will be necessary). 
The numerical assignment provided by the function $[[~.~]]_{hyp\_or\_lemma}$ for the lemmas of the libraries gives a big 
value spread (especially if ML4PG works with big proof libraries). We can tackle this problem assigning closer numbers to similar lemmas. We
are currently working in the implementation of these new features in ML4PG.

ML4PG can be combined with other tools to make the proof development easier. Search mechanisms implemented in Coq, such as the \lstinline?Search? 
command of SSReflect~\cite{SSReflect} and Whelp~\cite{AspertiGCTZ04}, can find patterns in lemma statements, but ML4PG can discover proof patterns
that cannot be found using existing Coq's search mechanisms, see \cite{HK12}. 

Moving towards symbolic machine-learning, we envisage that new lemmas or strategies could be conceptualised from the proof families obtained with
ML4PG using such techniques as Rippling~\cite{BB05} or Theory Exploration~\cite{JDB11}.

In addition, we plan to integrate more machine learning methods to help in the proof process.
To this aim, we need a tool which tracks not only the successful proofs, but also failed and discarded derivations
steps. In this way, we could use \emph{supervised} machine learning algorithms to indicate a user whether he is following
a sensible strategy based on the previous experience. Supervised machine learning could also be used to discover various proof styles. 

We are interested in increasing the number of proof assistants included in 
our framework. 
This will allow us to study proof similarities across 
different theorem provers. 
Since the interaction with theorem provers such as Isabelle or Lego is already available in Proof General,
we just need the implementation of the feature extraction mechanism for them; their interaction with the machine learning
engines would be the same as developed for Coq and SSReflect. 

Finally, current implementation of ML4PG is centralised; this means that the user can obtain proof
clusters of the libraries available on his computer. However, we think that a client-server architecture, where 
the proof information is shared among several users could also be  useful, especially for team-based program development. 

For this purpose, feature extraction in ML4PG is already designed to be lemma name and notation independent.

\section{Acknowledgements}\label{sec:ack}
We are grateful to the anonymous referees for their comments and suggestions; and the following individuals and research groups for inspiring discussions on the topics of machine learning in automated and interactive theorem proving:
J Strother Moore, members of the AI4FM project Alan Bundy, Cliff Jones, Leo Freitas, Ewen Maclean; and participants of Dagstuhl seminar 
\emph{AI meets Formal Software Development}~\cite{Dag}.

\bibliographystyle{eptcs}
\bibliography{mlipgii,katya2}

\end{document}